\documentclass[10pt,twocolumn,letterpaper]{article}

\usepackage[pagenumbers]{cvpr} % To force page numbers, e.g. for an arXiv version

\definecolor{cvprblue}{rgb}{0.21,0.49,0.74}
\usepackage[pagebackref,breaklinks,colorlinks,allcolors=cvprblue]{hyperref}

\usepackage{orcidlink}
\usepackage{xcolor}
\usepackage{kotex}
\usepackage{multirow}
\usepackage{multicol}
\usepackage{colortbl}
\usepackage{subcaption}
\usepackage[toc,page,header]{appendix}
\usepackage{minitoc}
\usepackage{placeins}

\definecolor{yellowgreen}{RGB}{154, 205, 50}

\title{Med-PerSAM: One-Shot Visual Prompt Tuning for Personalized Segment Anything Model in Medical Domain}

\author{
Hangyul Yoon\textsuperscript{1} \quad  Doohyuk Jang\textsuperscript{1} \quad Jungeun Kim\textsuperscript{1} \quad Eunho Yang\textsuperscript{1,2}\vspace{0.05in}\\
\textsuperscript{1}Korea Advanced Institute of Science and Technology (KAIST)\vspace{0.02in} \quad \textsuperscript{2}AITRICS}
\begin{document}
\doparttoc
\faketableofcontents

\maketitle
\begin{abstract}
Leveraging pre-trained models with tailored prompts for in-context learning has proven highly effective in NLP tasks. Building on this success, recent studies have applied a similar approach to the Segment Anything Model (SAM) within a ``one-shot" framework, where only a single reference image and its label are employed.  However, these methods face limitations in the medical domain, primarily due to SAM's essential requirement for visual prompts and the over-reliance on pixel similarity for generating them. This dependency may lead to (1) inaccurate prompt generation and (2) clustering of point prompts, resulting in suboptimal outcomes. To address these challenges, we introduce \textbf{Med-PerSAM}, a novel and straightforward one-shot framework designed for the medical domain. Med-PerSAM uses only visual prompt engineering and eliminates the need for additional training of the pretrained SAM or human intervention, owing to our novel automated prompt generation process. By integrating our lightweight warping-based prompt tuning model with SAM, we enable the extraction and iterative refinement of visual prompts, enhancing the performance of the pre-trained SAM. This advancement is particularly meaningful in the medical domain, where creating visual prompts poses notable challenges for individuals lacking medical expertise. Our model outperforms various foundational models and previous SAM-based approaches across diverse 2D medical imaging datasets. 
\end{abstract}   

\section{Introduction}
\label{sec:intro}

Medical image segmentation encompasses a broad range of clinical applications, from diagnostic assessments \cite{vidal2021multi} to treatment planning \cite{aggarwal2011role} and patient monitoring \cite{ouyang2020video}. Despite its critical importance, the task of segmentation by human labeler is exceptionally challenging, necessitating profound medical knowledge and validation by professionals. This complexity is further compounded by the subjective variability among practitioners and its labor-intensive nature \cite{jungo2018effect}. These challenges make it difficult to acquire a sufficient amount of labeled data, highlighting the need for medical image segmentation techniques that can perform well with a limited set of annotations.

Recently, several studies have explored the capability of few-shot learning for segmentation tasks by leveraging pre-trained foundation models, representing notable progress in addressing segmentation challenges 
 \cite{wang2023images, bar2022visual, zou2024segment, wang2023seggpt, butoi2023universeg, zhang2023personalize, liu2023matcher}. Drawing inspiration from the in-context learning capabilities of recent large language models \cite{openai2023gpt, wei2022chain, wei2022emergent, brown2020language, dong2022survey}, these approaches employ information from a small amount of labeled data to enhance the segmentation ability.

A notable instance of this progress is the Segment Anything Model (SAM) \cite{kirillov2023segment}. Through visual prompts such as points, bounding boxes, and masks that can serve as clues, SAM can enhance its performance. For instance, one can designate the positions of points that are considered similar or dissimilar to the target object as positive or negative point prompts, respectively~\cite{khirodkar2024harmony4d}. However, the dependence of the SAM on user-generated visual prompts highlights the persistent challenges in medical imaging, as creating optimal prompts requires users to have a deep understanding of medicine and anatomy.

\begin{figure*}[h]
  \centering
  \includegraphics[width=0.8\textwidth]{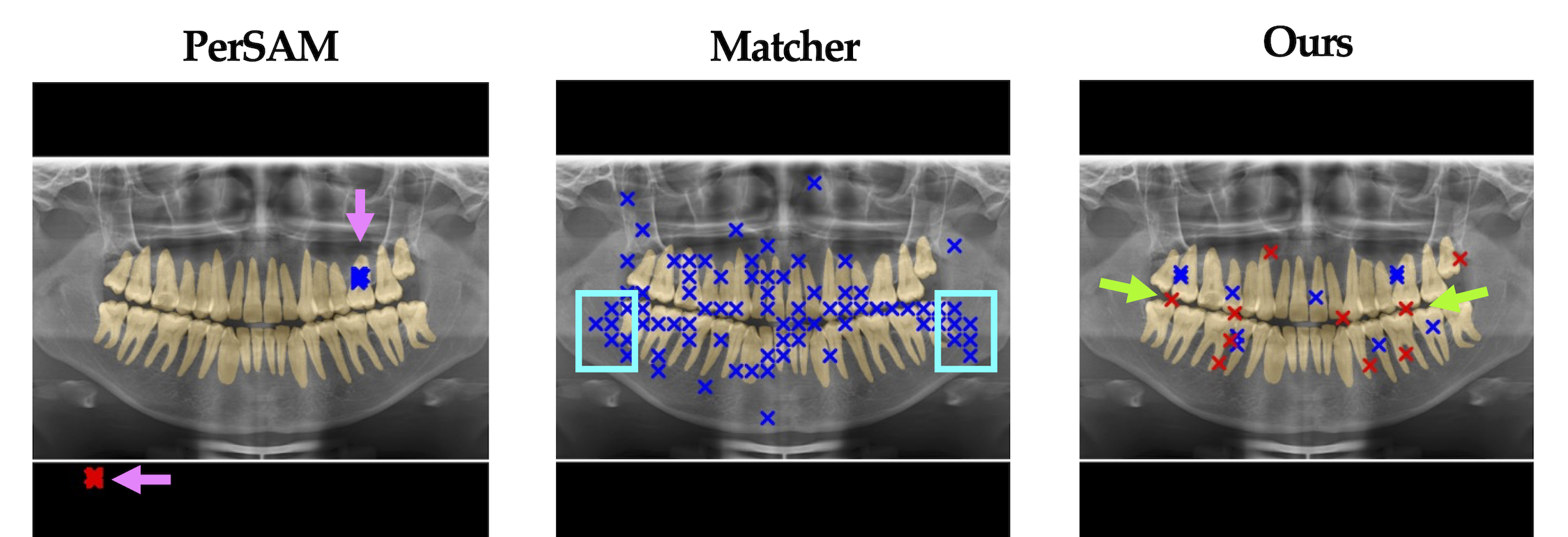}
  \vspace{-0.2cm}
  \caption{Comparison of point prompts from PerSAM \cite{zhang2023personalize}, Matcher \cite{liu2023matcher}, and our method. Ground truth masks are shown in yellow, with positive and negative point prompts depicted in blue and red, respectively. Unlike the two SAM-based methods, which show over-clustering (\textcolor{magenta}{pink arrows}) or struggle to differentiate organs with similar pixel intensities (\textcolor{cyan}{skyblue boxes}), our approach accurately generates prompts and also performs well in challenging areas (\textcolor{yellowgreen}{lime green arrows}). Additional examples can be found in Appendix \ref{appendix:qual}.}
  \label{fig:persam_prompt}
  \vspace{-0.5cm}
\end{figure*}

To address the difficulty of manually creating visual prompts in SAM, two recent studies, PerSAM \cite{zhang2023personalize} and Matcher \cite{liu2023matcher}, have made progress in the automated prompt generation in the ``one-shot'' setting. By employing a single reference image and its label, these approaches automatically generate visual prompts for test images, leveraging the capabilities of the pre-trained SAM without fine-tuning. Nonetheless, applying these methods in medical imaging poses challenges, primarily arising from the over-reliance on pixel-level similarity for prompt placement, which can result in generating inaccurate prompts (see Fig. \ref{fig:persam_prompt}). 

This challenge in medical imaging mainly stems from the predominance of grayscale images, which complicates object identification based solely on pixel values. The difficulty manifests in several ways: (1) the risk of misaligning point prompts with the target organ, a situation that may arise from the indistinguishable intensity levels between the target and surrounding organs (skyblue boxes in Fig. \ref{fig:persam_prompt}); (2) the tendency for point prompts to be clustered (pink arrows in Fig. \ref{fig:persam_prompt}); and (3) the common occurrence of negative prompts that concentrate on the image's black-padded edges, possibly resulting in inaccurate segmentation.

To transcend these limitations and increase SAM's potential in medical imaging, we introduce \textbf{Med-PerSAM}, a novel and straightforward approach for generating visual cues in medical images through a ``one-shot'' methodology. To address the limitations of previous studies, PerSAM and Matcher, our strategy incorporates image warping model to generate dense mask prompts. These mask prompts not only provide SAM a prior on the intrinsic anatomical characteristics of the given image but also form the basis for generating sparse (point and box) prompts. We also propose an visual prompt refinement strategy that improves SAM's performance in medical segmentation tasks without architectural modifications or further training of SAM. Lastly, we leverage the output of SAM as a pseudolabel to retrain the warping model. This process transfers the visual acuity of SAM to the warping model, producing more precise mask prompts and better segmentation.

In summary, our contribution is threefold:
\begin{itemize}
\item We introduce Med-PerSAM, a novel and straightforward framework for one-shot medical segmentation. Using just a single reference image and its corresponding mask, Med-PerSAM enables the customization of the SAM for particular medical datasets without additional fine-tuning.  

\item To the best of our knowledge, Med-PerSAM is the first model to propose an automatic visual prompt tuning strategy for SAM in the medical domain. It integrates a lightweight warping model for joint training and provides optimal visual prompts to SAM through an iterative refinement and retraining process.

\item Our method consistently demonstrates superior performance across a range of datasets, outperforming baseline models, including previous studies on one-shot SAM, such as PerSAM \cite{zhang2023personalize} and Matcher \cite{liu2023matcher}. By introducing a novel strategy that addresses the limitations in the visual prompt generation process of these two methods, we enhance SAM's performance in the medical domain.
\end{itemize}

\section{Related Works}
\label{sec:related_works}

\paragraph{\textbf{Image Warping}} 
Image warping is a technique used to manipulate or distort the given source image to achieve a desired shape or effect. Using the displacement field, it maps pixels from the original image to new locations in a transformed image, altering the appearance in various ways. It has evolved through deep learning and is used in various fields including video domain~\cite{park2023biformer, niklaus2018context, hu2022many, park2021asymmetric}, and is also frequently applied in medical applications~\cite{kim2024data,gu2021variational,zhang2021semi,chaitanya2021semi}.

Recently, several studies have applied optical flow-based warping as an effective deformable registration method for medical images from different patients \cite{balakrishnan2019voxelmorph,mansilla2020learning, jiang2024tumor, hering2022learn2reg}. This success is largely attributed to the standardized scan ranges and patient postures for each type of imaging acquisition. This ensures a certain level of consistency and similarity in anatomical features and organ positions, facilitating reliable image alignment. Building on these findings, we also utilize an image warping model for inter-patient transformation, enabling the reference image and mask pairs to be adapted to align with the semantics of the given test image.

\vspace{-0.3cm}

\paragraph{\textbf{Foundation Models for Segmentation}} Pre-trained foundation models, known for their robust generalization capabilities, have demonstrated significant adaptability across a variety of downstream applications, yielding impressive outcomes. Particularly in natural language processing, models such as GPT~\cite{radford2018gpt1,radford2019gpt2,brown2020gpt3,achiam2023gpt4} and LLaMA~\cite{touvron2023llama,touvron2023llama2} series have shown outstanding in-context learning abilities. These models can be applied to new tasks using domain-specific prompts, which is a testament to their versatility.

In the field of computer vision, pre-trained foundation models have proven their performance in image segmentation. Models like SAM~\cite{kirillov2023segment}, Painter~\cite{wang2023images}, and SegGPT~\cite{wang2023seggpt}, demonstrate excellent generalization across various image datasets. These pre-trained models effectively handle various segmentation tasks without the need for task-specific training, demonstrating their versatility and contributing to advancements in the field of image segmentation.

\vspace{-0.4cm}
\paragraph{\textbf{Segment Anything Model}} SAM, a segmentation model pre-trained on a dataset with 1 billion masks and 11 million images \cite{kirillov2023segment}, uses visual prompts alongside images to generate masks, employing a pre-trained vision transformer and a lightweight encoder-decoder architecture. However, it has been reported that SAM may produce suboptimal results when given medical images as input \cite{deng2023segment,hu2023sam,zhou2023sam,he2023computervision}.

In response, several recent studies, including MedSAM~\cite{ma2024segment}, ProtoSAM~\cite{ayzenberg2024protosam}, CAT-SAM~\cite{catsam}, and others~\cite{wu2023medical, wong2023scribbleprompt,zhang2023customized, wei2024prompting}, have focused on adapting SAM to the medical domain. However, these studies require a dataset for additional fine-tuning of SAM or assume the availability of ground truth visual prompts during inference. This assumption deviates from real-world clinical settings, where obtaining a large dataset for fine-tuning can be challenging, and it is not guaranteed that users will be able to provide an exact ground truth visual prompt for medical images. In contrast to these studies, to the best of our knowledge, we are the first to propose a training-free SAM scheme that does not require or assume the presence of a manual visual prompt during inference in the medical domain.

\vspace{-0.4cm}

\paragraph{\textbf{One-Shot Segment Anything Model}} PerSAM \cite{zhang2023personalize} and Matcher \cite{liu2023matcher} are two notable studies focused on one-shot segmentation using SAM. Both methods perform inference on unlabeled images with a pretrained SAM, utilizing only a single reference image and mask, along with automated visual prompts. PerSAM uses feature similarity of the SAM visual encoder for creating point and box prompts, and Matcher uses DINOv2 \cite{oquab2023dinov2} for bidirectional and instance-level matching to extract positive point offsets. However, as shown in Fig. \ref{fig:persam_prompt}, both models often misplace point prompts in medical images due to over-reliance on pixel and feature similarities without considering contextual factors.

\begin{figure*}[t]
  \centering
  \includegraphics[width=0.87\textwidth]{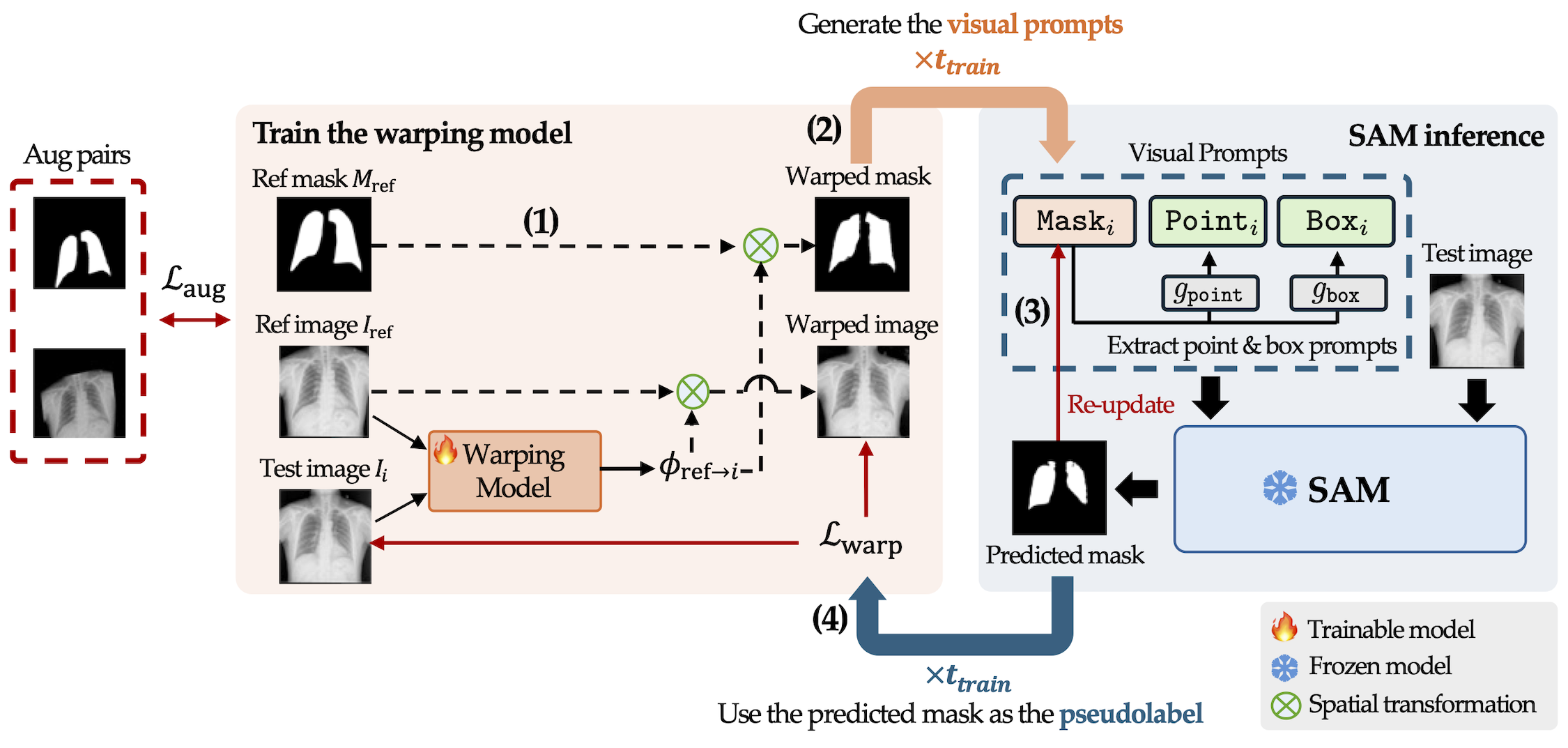}
  \vspace{-0.3cm}
  \caption{Overall Framework of Med-PerSAM. (1) Initially, the warping model is trained with warping ($\mathcal{L}_\text{warp}$) and augmentation ($\mathcal{L}_\text{aug}$) losses, and employs optical flow to generate a warped mask. (2) This serves as a mask prompt for SAM, while point and box prompts are extracted from the modules \(g_{\texttt{point}}\) and \(g_{\texttt{box}}\), respectively. (3) The resulting output is used to update the visual prompts and SAM output, (4) and the refined prediction mask is again utilized to retrain the warping model, which enhances the quality of the warped mask.}
  \label{fig:main_framework}
  \vspace{-0.6cm}
\end{figure*}

\vspace{-0.1cm}
\section{Methods}
\vspace{-0.1cm}
\label{sec:methods}
The overall architecture of Med-PerSAM is illustrated in Fig. \ref{fig:main_framework}. Given a single reference image and its mask, along with a test dataset containing unlabeled images, the main goal is to optimize the visual prompt sets for SAM to produce the final prediction masks. To achieve this, our method follows an iterative process that involves (1) training the warping model (Section \ref{method:warp}), (2) generating visual prompts and running SAM inference (Section \ref{method:prompt generation}), and (3) retraining the warping model (Section \ref{method:retrain}). The warping model provides visual cues to SAM, whose outputs are then used to retrain and enhance the warping model, creating an iterative feedback loop that improves the performance of both models. While our method is described under the assumption of single-class segmentation for clarity, extending it to multiclass segmentation is straightforward (Section \ref{method:multi_class}).

\subsection{Initial Training of Warping Model}
\label{method:warp}
Initially, Med-PerSAM trains the warping model based on two loss terms, and the main purpose is to generate the warped masks with the trained model (Eq. \ref{eq:initial_warp_mask}), which is conveyed to SAM as a visual prompt in Section \ref{method:prompt generation}.

The warping model \(f^{\theta}\) with learnable parameter \(\theta\), is trained to identify the optical flow to transform a reference image \(I_{\text{ref}}\in \mathbb{R}^{h \times w}\) to a set of unlabeled test images \(\{I_i\}_{i=1}^{N} \in \mathbb{R}^{h \times w} \). The training takes place between the reference sample and the test set, similar to the setting of test-time training/adaptation~\cite{zhu2021test, balakrishnan2019voxelmorph}. After training, it applies the warping transformation on the reference mask \(M_{\text{ref}}\) to produce mask prompts for each of the unlabeled images.

For the given test image \(I_i\), the model calculates the optical flow \(\phi_{\text{ref} \rightarrow i} = f^{\theta}(I_{\text{ref}}, I_i)\) between the reference and test images. This flow facilitates the transformation of the reference image \(I_{\text{ref}}\) into a warped image \(\hat{I}_{\text{ref} \rightarrow i} = I_{\text{ref}} \circ \phi_{\text{ref} \rightarrow i}\), aiming to make it closely resemble \(I_i\). Here, \(\circ\) represents the operation of spatial transformation. 

The primary objective of training the warping model is to minimize the warping loss \(\mathcal{L}_{\text{warp}}\), to maximize the similarities between the warped and target images. This loss function integrates two components: (1) the image loss \(\mathcal{L}_{\text{img}}\), which consists of similarity-based functions such as the structural similarity index measure (SSIM)~\cite{terpstra2022loss} or normalized cross-correlation (NCC) loss \cite{shu2021medical}, and (2) the L2 flow regularization loss, \(\mathcal{L}_{\text{reg}}\), which promotes smoothness in the flow field \cite{meng2023non, meng2024correlation}. The warping loss between the reference and target images is formulated as 
\vspace{-0.1cm}
\begin{align}
\mathcal{L}_{\text{warp}}(I_\text{ref}, I_i) = \mathcal{L}_{\text{img}}(\hat I_{\text{ref} \rightarrow i}, I_{i}) + \mathcal{L}_{\text{reg}} 
 (\phi_{\text{ref} \rightarrow i}). \label{eq:warp_img}
\end{align}

With the warping loss, an augmentation loss \(\mathcal{L}_\text{aug}\) is introduced to enhance training and capture large photometric and geometric differences during the warping process. By applying augmentation to the reference pair, augmented image \( I_\text{aug} \) and mask \( M_\text{aug} \) are created. The warping loss between the reference and augmented images is then computed, which is analogous to Eq. \ref{eq:warp_img}, but with a label-to-label loss between the augmented and warped masks:
\begin{align}
\mathcal{L}_{\text{aug}} = \mathcal{L}_{\text{warp}}(I_{\text{ref}}, I_{\text{aug}}) + \mathcal{L}_{\text{seg}}( \hat M_{\text{ref} \rightarrow \text{aug}}, M_{\text{aug}}).
 \label{eq:warp_aug}
\end{align}
Here, $\mathcal{L}_\text{seg}$ represents the segmentation loss function, for which we used DiceCE, a combination of DICE \cite{milletari2016v} and cross-entropy losses. Even if there is a significant difference between the reference and augmented images, the warping model can successfully learn the transformation with guidance from segmentation loss between the warped augmentation mask \( \hat M_{\text{ref} \rightarrow \text{aug}} = M_{\text{ref}} \circ \phi_{\text{ref} \rightarrow \text{aug}} \) and \( M_{\text{aug}} \).

Therefore, the total loss $\mathcal{L}_\text{train}$ for the initial training is the sum of the warping and the augmentation losses, as described in Eqs. \ref{eq:warp_img} and \ref{eq:warp_aug}, which is denoted by
\vspace{-0.1cm}
\begin{align}
\mathcal{L}_\text{train} = \mathcal{L}_{\text{warp}}(I_\text{ref}, I_i) + \mathcal{L}_\text{aug}.
\label{eq:initial_train_loss}
\end{align}

\vspace{-0.1cm}
Upon training the warping model, the warped mask can be derived from the reference mask as follows: 
\begin{align}
\hat{M}_{\text{ref} \rightarrow i} = M_{\text{ref}} \circ \phi_{\text{ref} \rightarrow i}. \label{eq:initial_warp_mask}
\end{align}

\vspace{-0.2cm}
\subsection{Visual Prompt Generation}
\label{method:prompt generation}
\begin{figure}[h]
  \centering
  \includegraphics[width=0.98\columnwidth]{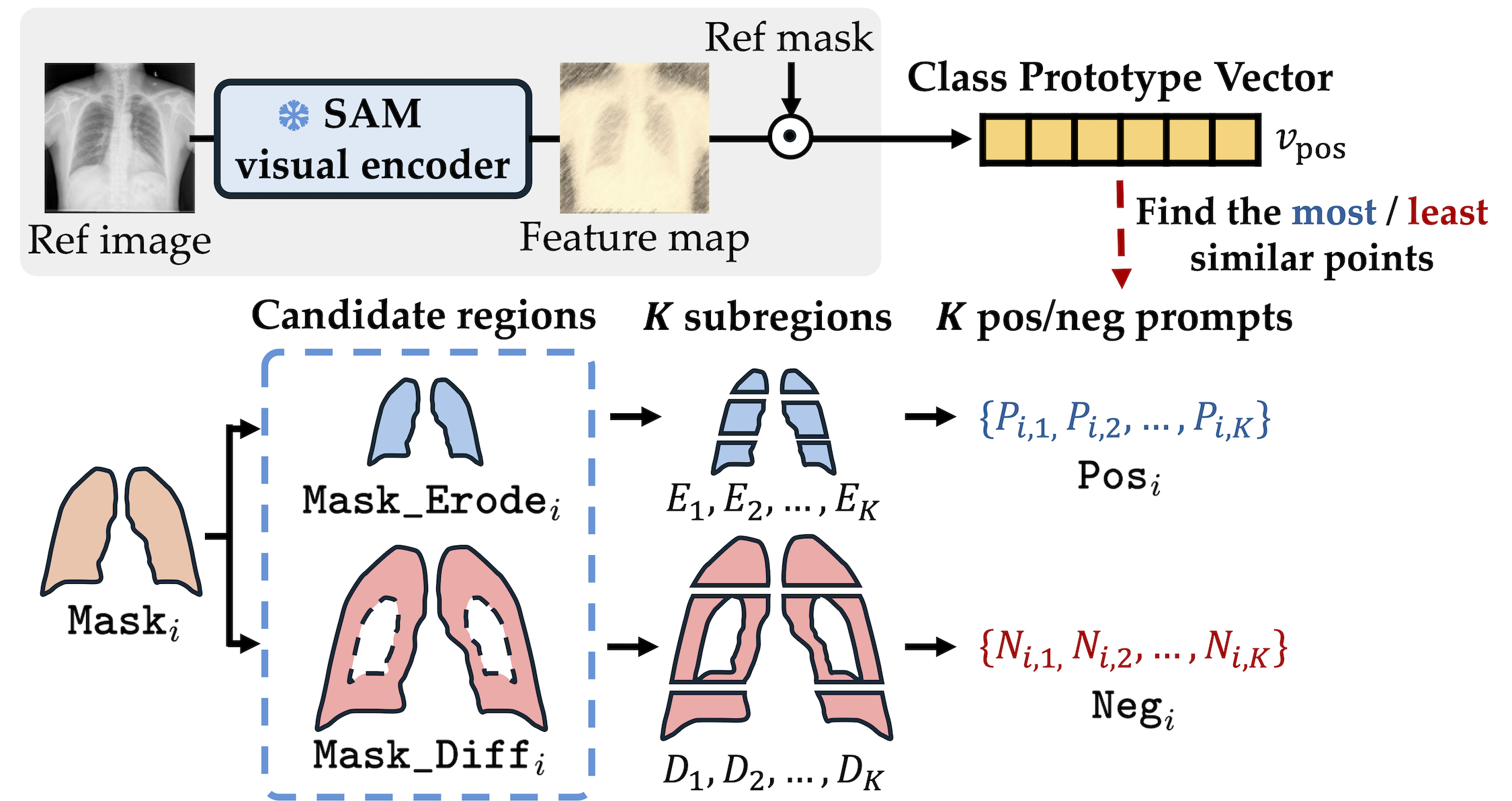}
  \vspace{-0.2cm}
  \caption{Overview of our point prompting strategy. (1) A prototype vector is defined as the average of foreground feature vectors from the reference image. (2) Erosion and dilation kernels are used to identify candidate regions for point prompts, which are then divided into subregions. (3) In each subregion, positive and negative point prompts are chosen based on the offsets that display the highest and lowest similarity to the class prototype vector, respectively.}
  \label{fig:point_prompt}
  \vspace{-0.6cm}
\end{figure}

\vspace{-0.1cm}
From the warped mask obtained in the above Eq. \ref{eq:initial_warp_mask}, we can acquire visual prompts and iteratively update it to improve the outcomes. The warped mask serves as the mask prompt \( \texttt{Mask}_i\) for the test image \(I_{i}\) and is integrated with it in the SAM. However, in SAM, the mask prompt is designed to complement point/box prompts, and using the mask prompt alone has been reported to cause malfunctions\footnote{https://github.com/facebookresearch/segment-anything/issues/169}. Therefore, we also generate point and box prompts by (1) calculating the similarity map and (2) deriving the point prompt using our proposed strategy, as illustrated in Fig. \ref{fig:point_prompt}.

To delineate candidate regions for point prompt extraction, we apply morphological operations—erosion and dilation—on the mask prompt using predefined kernels \cite{narayanan2006fast, bradski2000opencv}. These operations adjust each pixel's value based on the minimum or maximum values in its local vicinity, thereby modifying the mask's size while preserving its structural integrity. This can be expressed as
\vspace{-0.1cm}
\begin{align}
& \texttt{Mask\_Erode}_i = \text{Erosion}(\texttt{Mask}_i, K_e), \label{eq:mask_erosion} \\
& \texttt{Mask\_Dilate}_i = \text{Dilation}(\texttt{Mask}_i, K_d), \\
& \texttt{Mask\_Diff}_i = \texttt{Mask\_Dilate}_i - \texttt{Mask\_Erode}_i, 
\end{align}
where Erosion and Dilation denotes the morphological operations with the kernels \( K_e \) and \( K_d \), respectively.

Subsequently, we extract the positive prompts from $\texttt{Mask\_Erode}_i$ and the negative prompts from $\texttt{Mask\_Diff}_i$. The objective of this approach is to accurately identify positive prompts from regions of high certainty and designate the negative prompt from ``hard negative'' \cite{robinson2020contrastive, gao2021simcse}, indicating negatives that are challenging to predict. We divide each candidate region into subregions, extracting a single prompt from each based on the highest and lowest similarity values. The subregions are defined by sorting the pixel positions in ascending order according to their index numbers and then dividing them into equal segments. Then, we obtain the class prototype vector as an average foreground feature from the SAM encoder. With the cosine similarity map $S_i \in [-1, 1]^{h \times w}$ between the prototype vector and the SAM encoder feature of the test image, the \( K \) number of positive and negative prompts, denoted as $\texttt{Pos}_i$ and $\texttt{Neg}_i$, are defined as
\begin{align}
\texttt{Pos}_i = \{ \underset{(x, y) \in E_k}{ \arg\max} \, S_i(x, y) \,|\, k = 1, 2, \dots, K\}, \\
\texttt{Neg}_i = \{ \underset{(x, y) \in D_k}{ \arg\min} \, S_i(x, y) \,|\, k = 1, 2, \dots, K\},
\end{align}
where \( E_k \subseteq \texttt{Mask\_Erode}_i \) and \( D_k \subseteq \texttt{Mask\_Diff}_i \) denote the \( k \)-th subregions within the eroded mask and the difference area of the eroded and dilated masks, respectively. The point prompt extraction is then formulated as
\vspace{-0.1cm}
\begin{align}
\texttt{Point}_i = g_{\texttt{point}}(\texttt{Mask}_i) = (\texttt{Pos}_i,\, \texttt{Neg}_i).    
\end{align}

The process of defining box prompt $\texttt{Box}_i$ is relatively simple. The function $g_{\texttt{box}}$ extracts the coordinates of the four vertices of the minimal rectangle that encompasses all positive pixels from the given mask prompt $\texttt{Mask}_i$. 

By leveraging the obtained point, box, and mask prompts along with the input image in SAM, we predict the segmentation mask $\hat{M}_i$ as follows:
\begin{align}
\texttt{Prompt}_i & = (\texttt{Point}_i, \; \texttt{Box}_i, \; \texttt{Mask}_i), \\
\hat{M}_i &= \textbf{SAM}(\texttt{Prompt}_i, I_i), \; \label{eq:mask_pred_initial}
\end{align}
where \textbf{SAM} denotes the pretrained SAM~\cite{kirillov2023segment}.

\begin{figure}[h]
  \vspace{-0.3cm}
  \centering
  \includegraphics[width=0.98\columnwidth]{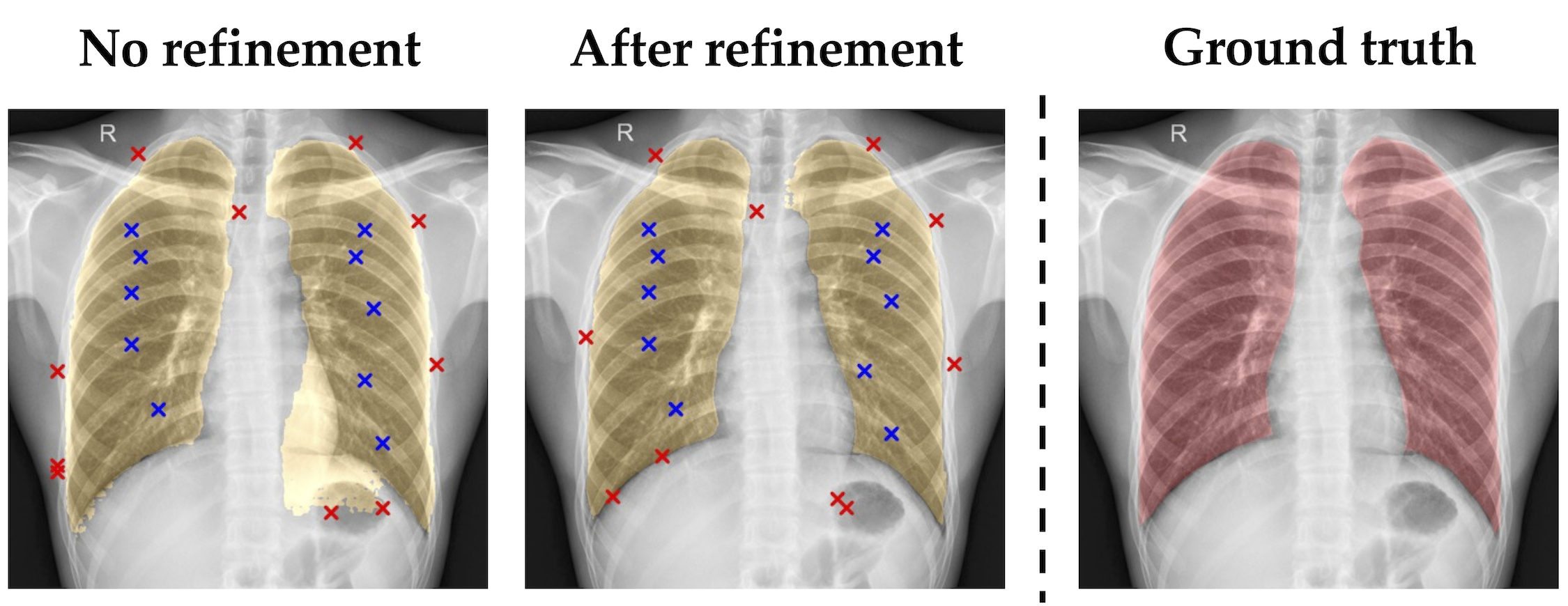}
  \vspace{-0.3cm}
  \caption{Example of prompt refinement. The predicted outcome is indicated in yellow, while the positive and negative point prompts are marked in blue and red, respectively. Additional examples will be provided in Appendix~\ref{appendix:qual}.}
  \label{fig:iterative_refine}
  \vspace{-0.5cm}
\end{figure}

Furthermore, we improve the model's performance by refining the point and box prompts based on the output from the SAM, which serves as a new mask prompt. This refinement process involves the repeated application of the procedures detailed in Eq. \ref{eq:mask_erosion} to \ref{eq:mask_pred_initial}. Through this iterative process, the prompts are refined by leveraging the output from the previous iteration, progressively improving the accuracy of the segmentation mask prediction (Fig. \ref{fig:iterative_refine}). 

\subsection{Retraining the Warping Model}
\label{method:retrain}
We retrain the warping model by integrating a label-to-label loss derived from the predicted masks with the previously defined training loss to enhance warping efficiency. This process involves using the predicted mask as a pseudolabel and subsequently employing the model's output as the mask prompt for the SAM. The retraining loss \( \mathcal{L}_{\text{retrain}} \) is calculated similarly to \( \mathcal{L}_{\text{train}} \) as defined in Eq. \ref{eq:initial_train_loss} of Section \ref{method:warp}, but it includes an additional segmentation loss term between the warped mask and pseudolabel. This can be expressed as
\begin{align}
\mathcal{L}_{\text{retrain}} = \mathcal{L}_{\text{train}} + \mathcal{L}_{\text{seg}}(\hat{M}_{\text{ref} \rightarrow i}, \hat{M}_{i}), \nonumber
\end{align}
where \( \mathcal{L}_{\text{seg}} \) is the previously defined DiceCE loss function.

We can once again obtain the visual prompt from the retrained warping model. By iterating the processes described in Section \ref{method:prompt generation}, we can update the prediction outcomes, and repeatedly iterate the retraining as many times as desired.

\begin{table*}[t]
\caption{Segmentation results (\%) of  our and other baseline methods. The highest and second-highest scores are indicated by \textbf{bold} and \underline{underlined} text, respectively. The superscript $\dag$ signifies the use of the pretrained SAM, which is our primary focus for comparison.}
\vspace{-0.2cm}
\footnotesize
\setlength{\tabcolsep}{2.5pt}
\renewcommand{\arraystretch}{1}
\centering
\begin{tabular}{c|c|cccccccccc}
\toprule
\multirow{2}{*}{\textbf{Method}} & \multirow{2}{*}{\textbf{Venue}}
 & \multicolumn{2}{c}{\textbf{Shenzhen}}&\multicolumn{2}{c}{\textbf{OdontoAI}}&\multicolumn{2}{c}{\textbf{CAMUS}}&\multicolumn{2}{c}{\textbf{JSRT}} &\multicolumn{2}{c}{\textbf{BUU}} \\ 
 & & mIoU & DICE & mIoU & DICE & mIoU & DICE & mIoU & DICE & mIoU & DICE \\
\midrule
Painter \cite{wang2023images} & CVPR 2023 & 28.3 & 43.7 & 25.3 & 39.2 & 24.9 & 39.6 & 42.6 & 59.4 & 4.0 & 7.3 \\
VP \cite{bar2022visual} & NeurIPS 2022 & 69.6 & 81.8 & 50.4 & 66.8 & 29.9 & 44.9 & 27.0 & 39.3 & 18.2 & 29.9 \\
SEEM \cite{zou2024segment} & NeurIPS 2023 & 32.6 & 48.6 & 29.4 & 45.1 & 24.8 & 39.4 & 58.3 & 73.3 & 2.2 & 3.9 \\
SegGPT \cite{wang2023seggpt} & ICCV 2023 & \underline{80.6} & \underline{89.2} & \underline{71.4} & \underline{83.3} & \underline{39.6} & \underline{54.9} & 25.6 & 37.8 & \underline{57.6} & \underline{71.8} \\
UniverSeg \cite{butoi2023universeg} & ICCV 2023 & 43.8 & 60.6 & 43.2 & 60.1 & 31.0 & 46.2 & \underline{61.7} & \underline{75.7} & 24.3 & 38.4 \\
PerSAM$^{\dag}$ \cite{zhang2023personalize} & ICLR 2024 & 35.5 & 51.0 & 17.2 & 28.9 & 27.3 & 42.9 & 41.7 & 57.7 & 17.9 & 27.5 \\
PerSAM-F$^{\dag}$ \cite{zhang2023personalize} & ICLR 2024 & 29.0 & 44.6 & 16.1 & 27.4 & 30.1 & 45.6 & 38.7 & 53.8 & 5.5 & 10.4 \\
Matcher$^{\dag}$ \cite{liu2023matcher} & ICLR 2024 & 29.4 & 45.4 & 13.1 & 23.2 & 11.6 & 20.8 & 38.6 & 52.4 & 4.9 & 9.4 \\
\midrule
\rowcolor{gray!15} \textbf{Med-PerSAM} & This work & \textbf{85.4} & \textbf{92.0} & \textbf{78.3} & \textbf{87.8} & \textbf{60.9} & \textbf{74.2} & \textbf{84.2} & \textbf{91.3} & \textbf{58.0} & \textbf{72.9} \\
\bottomrule
\end{tabular}
\label{tab:main result}
\vspace{-0.5cm}
\end{table*}

\subsection{Extension to Multi-Class Segmentation}
\label{method:multi_class}
In Sections \ref{method:warp} to \ref{method:retrain}, we elaborate on the methodology assuming a single-class segmentation. Extending this to a multi-class scenario is straightforward. For multi-class mask prompts, a one-hot mask is generated for each class except the background, and separate mask prompts are created. These prompts are input into SAM alongside the test image for prediction. The foreground class with the highest prediction logit is assigned to each pixel, while pixels with all foreground logits below 0 are labeled as background.

\section{Experiments}
\label{sec:experiments}

\subsection{Datasets}
The experiments are conducted on five benchmark datasets, including various anatomical regions. The Shenzhen dataset \cite{jaeger2014two} is utilized as a source for lung segmentation. The OdontoAI dataset \cite{silva2022odontoai} comprises teeth segmentation data. The JSRT and CAMUS \cite{leclerc2019deep} datasets \cite{shiraishi2000development, gaggion2022improving} are multi-class  segmentation in chest and heart imagings. In addition, the BUU dataset \cite{klinwichit2023buu} for spine segmentation is used. Detailed descriptions of datasets are in the Appendix \ref{appendix:dataset}.

\vspace{-0.1cm}
\subsection{Baseline Models}
We selected various foundational models, including Painter \cite{wang2023images}, Visual Prompting (VP) \cite{bar2022visual}, SEEM \cite{zou2024segment}, SegGPT \cite{wang2023seggpt}, UniverSeg \cite{butoi2023universeg}, PerSAM \cite{zhang2023personalize}, and Matcher \cite{liu2023matcher}, as baselines for evaluation. Additionally, we tested PerSAM-F, a fine-tuned version of PerSAM. 

\vspace{-0.1cm}
\subsection{Implementation Details}
\label{sec:implementation_detail}
Following PerSAM~\cite{zhang2023personalize}, the reference image and its corresponding label are identified by arranging the names of the samples within each dataset alphabetically and selecting the first sample in this order as the reference. The remaining samples are then utilized as the test set for inference.

Our warping model is based on NICE-Trans~\cite{meng2023non}. To enhance robustness against large geometric transformations, such as scaling and rotation, the model incorporates affine registration, followed by a deformable registration field for final warping. Since NICE-Trans was originally designed for 3D imaging, we modified it for adaptation to 2D images. The warping model has around 19 million parameters, which is only about 3\% of SAM's 641 million parameters. The numbers for retraining and prompt refinement are fixed at 5 and 1 in the main experiment, respectively. We used SAM version 1 for a fair comparison with previous SAM-based studies, and replacing it with SAM2 yields similar performance (see Appendix ~\ref{appendix:warp} for more details).

Both point and box prompts are used, based on the findings in PerSAM~\cite{zhang2023personalize}. For point prompts, 10 positive and negative offsets are used for the Shenzhen and OdontoAI datasets, while 5 points each are used for the other datasets. More details about the model, loss function, and training process can be found in Appendix \ref{appendix:warp}. Other experimental details, such as morphological operations and an analysis on the point prompt numbers, are provided in Appendix \ref{appendix:experiment}.

For the evaluation, we utilize the Intersection over Union (IoU) and DICE scores. The IoU is calculated by dividing the area of overlap between the prediction and the ground truth by the area of their union. Similarly, the DICE score \cite{dice1945measures} is calculated as twice the area of overlap divided by the sum of the areas of the two masks. 
For multi-class segmentation, the metric values are computed separately for each foreground class and averaged across all the classes.

\begin{figure*}[t]
  \centering
    \includegraphics[width=0.9\textwidth]{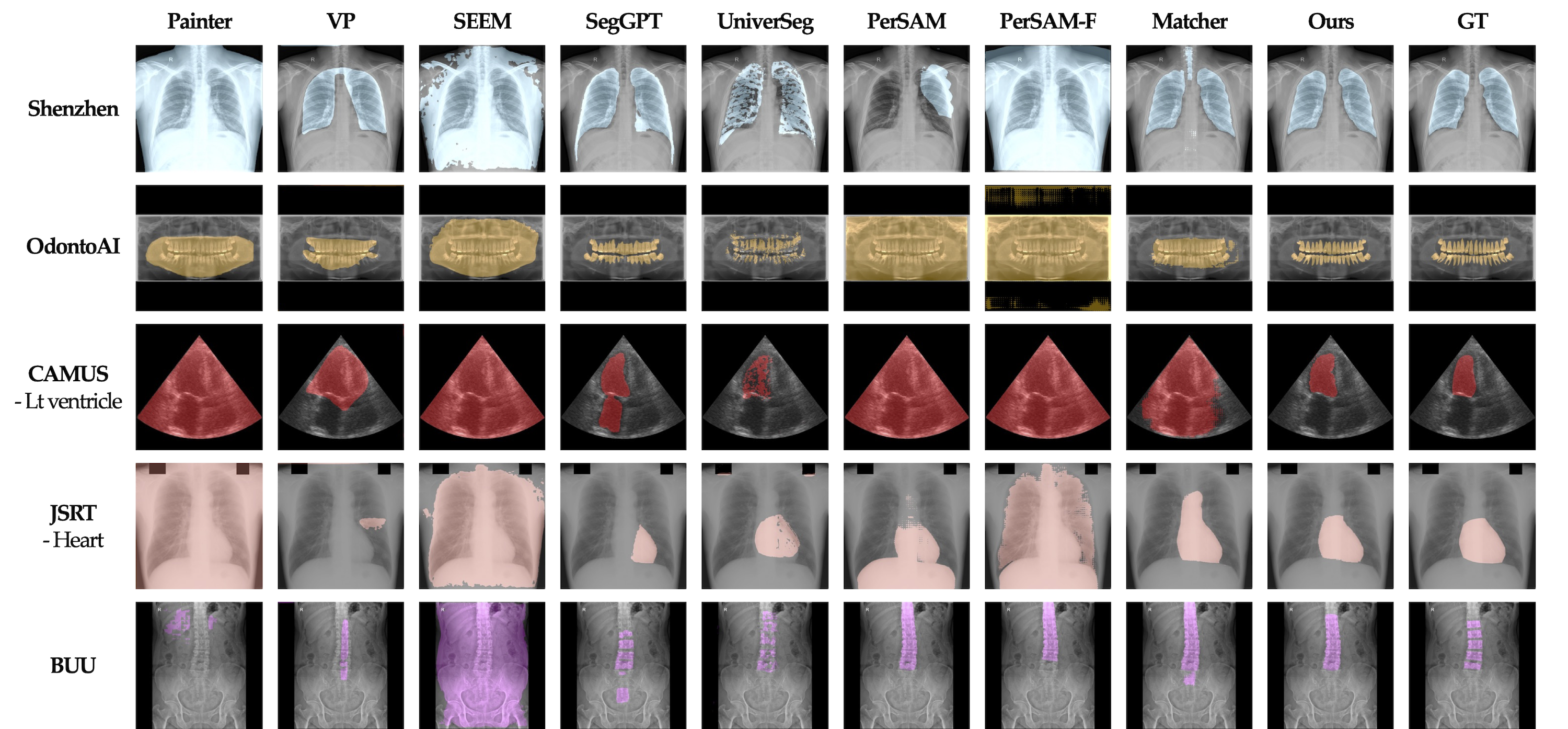}
  \vspace{-0.2cm}
  \caption{A qualitative comparison of the results from our model and other baseline models. Additional visualized examples of the main experiment and ablation studies will be provided in the Appendix \ref{appendix:qual}.}
  \label{fig:qual_analysis}
  \vspace{-0.55cm}
\end{figure*}

\subsection{Main Results}
The main experimental results are summarized in Table \ref{tab:main result}. Our model demonstrates superior performance with a significant margin compared to other foundational models across all datasets. Notably, our model signficantly outperforms two previous studies that employed SAM (PerSAM, Matcher), with the performance gain up to 65\%. Furthermore, these two studies often demonstrate signficantly inferior performance relative to other baseline models, indicating that utilizing SAM without appropriate prompts may lead to a substantial decline in performance. 

To ensure the robustness of our method, we also get the averaged results from ten different reference samples, which are shown in Table \ref{tab:main_result_avg}. It can be seen that our model still achieves the best performance across all datasets.

\begin{table}[h]
\footnotesize
\caption{The average and standard error of DICE scores (\%) for ten different reference samples. $\dag$ indicates the use of SAM.}
\vspace{-0.2cm}
\setlength{\tabcolsep}{2.5pt}
\renewcommand{\arraystretch}{1}
\centering
\begin{tabular}{c|ccccc}
\toprule 
\textbf{Method} & \textbf{Shenzhen}& \textbf{OdontoAI} & \textbf{CAMUS} & \textbf{JSRT} & \textbf{BUU} \\ 
\midrule
Painter & 43.7$\pm$0.0 & 39.4$\pm$0.0 & 39.6$\pm$0.0 & 57.8$\pm$0.5 & 7.2$\pm$0.0  \\
VP & 81.1$\pm$0.5 & 63.9$\pm$0.8 & 48.5$\pm$1.2 & 39.2$\pm$0.8 & 22.6$\pm$3.0 \\
SEEM & 48.5$\pm$0.0 & 46.1$\pm$0.1 & 39.4$\pm$0.0 & \underline{73.7$\pm$0.1} & 3.9$\pm$0.0 \\
SegGPT & \underline{90.9$\pm$0.1} & \underline{82.9$\pm$0.0} & \underline{52.0$\pm$0.6} & 41.8$\pm$0.4 & \underline{66.9$\pm$0.8} \\
UniverSeg & 59.3$\pm$1.5 & 54.0$\pm$1.6 & 40.2$\pm$2.2 & 73.1$\pm$1.1 & 31.2$\pm$2.6 \\
PerSAM$^\dag$ & 52.3$\pm$0.7 & 29.2$\pm$0.2 & 40.5$\pm$2.8 & 53.8$\pm$1.4 & 22.1$\pm$2.0 \\
PerSAM-F$^\dag$ & 45.0$\pm$0.2 & 30.5$\pm$2.5 & 48.2$\pm$2.4 & 52.8$\pm$1.0 & 29.2$\pm$4.0 \\
Matcher$^\dag$ & 53.1$\pm$0.7 & 25.8$\pm$1.0 & 28.8$\pm$0.7 & 20.6$\pm$0.3 & 11.0$\pm$0.6 \\ 
\midrule
\rowcolor{gray!15} \textbf{Med-PerSAM} & \textbf{91.7$\pm$0.8} & \textbf{87.0$\pm$0.4} & \textbf{64.4$\pm$2.3} & \textbf{90.5$\pm$0.7} & \textbf{69.4$\pm$1.5} \\ 
\bottomrule
\end{tabular}
\label{tab:main_result_avg}
\vspace{-0.6cm}
\end{table}

The examples of qualitative analysis are illustrated in Fig. \ref{fig:qual_analysis}. It has been observed that most foundation models struggle to accurately segment medical images. Specifically, these models often perform segmentation at a larger scale rather than focusing on the expected parts of the image, or they may include other organs with similar intensity values in the prediction. In contrast, our model model consistently achieves reliable results across various datasets.

\vspace{-0.3cm}
\subsection{Analyses of Point Prompts}
Table \ref{tab:point_ratio} presents the proportions of accurately identified positive and negative point prompts across three models utilizing SAM. The results demonstrate that our prompting strategy finds highly accurate positive and negative point prompts. Although PerSAM also exhibits high accuracy with point prompts, these prompts are often highly concentrated in certain regions, as depicted in Fig. \ref{fig:persam_prompt}.

This clustering issue is also observed in Table \ref{tab:point_cluster}, which demonstrates the clustering tendency of point prompts using Hopkins' statistic \cite{lawson1990new}. A lower value indicates less clustering, and PerSAM shows a significantly higher value compared to our method. Combined with the ablation study results of our prompting strategy, which will be described in Table \ref{tab:prompt_ablation}, this suggests that such point clustering can lead to a decline in SAM's performance.

\begingroup
\renewcommand{\arraystretch}{1}
\begin{table}[h]
\caption{The ratio (\%) of positive point prompts (PP) within ground truth mask and negative point prompts (NP) within the background for each method that utilizes point prompts.}
\vspace{-0.15cm}
\footnotesize
\setlength{\tabcolsep}{5pt}
\centering
\begin{tabular}{l|cccccc}
\toprule
\multirow{2}{*}{\textbf{Method}} 
 & \multicolumn{3}{c}{\textbf{Shenzhen}} & \multicolumn{3}{c}{\textbf{OdontoAI}} \\
 & PP & NP & Total & PP & NP & Total \\
\midrule
PerSAM & \textbf{99.7} & 97.6 & 98.6 & 96.8 & \textbf{100.0} & 98.4 \\
Matcher & 64.8 & - & 64.8 & 48.6 & - & 48.6 \\
\rowcolor{gray!15} \textbf{Ours} & 99.6 & \textbf{99.3} & \textbf{99.4} & \textbf{99.7} & 97.9 & \textbf{98.8} \\
\bottomrule
\end{tabular}
\label{tab:point_ratio}
\vspace{-0.2cm}
\end{table}
\endgroup

\begin{table}[h]
\caption{Hopkin's statistics for PerSAM and our method. Matcher is excluded due to different point numbers.}
\vspace{-0.2cm}
\footnotesize
\renewcommand{\arraystretch}{1}
\centering
% \begin{threeparttable}
\begin{tabular}{l|cccc}
\toprule
\multirow{2}{*}{ \textbf{Method} }
 & \multicolumn{2}{c}{\textbf{Shenzhen}}&\multicolumn{2}{c}{\textbf{OdontoAI}} \\ 
 &  PP & NP & PP & NP \\ 
\midrule
PerSAM & 0.9968 & 0.9966 & 0.9951 & 0.9964 \\ 
\rowcolor{gray!15}  \textbf{Ours} & \textbf{0.6709} & \textbf{0.5906} & \textbf{0.8262} & \textbf{0.7898} \\ 
\bottomrule
\end{tabular}
\vspace{-0.5cm}
\label{tab:point_cluster}
\end{table}

\subsection{Robustness to Image Characteristics}
\label{sec:image_robustness}
Even for the same type of imaging, the characteristics of medical images can vary due to differences in equipment or institutions, potentially affecting model performance~\cite{guan2022domain, pooch2020can}. To account for this issue, we evaluated cross-dataset performance using the Shenzhen and JSRT chest X-ray datasets, where the reference and test samples were taken from different datasets. Reference labels were modified by a medical doctor to ensure consistency between datasets. As shown in Table~\ref{tab:cross_dataset}, our model's performance remains consistent despite changes in the reference samples.

\begin{table}[h]
\vspace{-0.3cm}
\caption{The cross-dataset performance. `Changed' indicates altering the reference sample to one from a different dataset.}
\vspace{-0.2cm}
\footnotesize
\renewcommand{\arraystretch}{1}
\centering
% \begin{threeparttable}
\begin{tabular}{c|cccc}
\toprule
\multirow{2}{*}{ \textbf{Reference} }
 & \multicolumn{2}{c}{\textbf{Shenzhen}}&\multicolumn{2}{c}{\textbf{JSRT}} \\ 
 & mIoU & DICE & mIoU & DICE \\ 
\midrule
\textbf{Original} & 85.4 & 92.0 & 84.2 & 91.3 \\ 
\textbf{Changed} & 86.6 & 92.6 & 84.0 & 91.1 \\ 
\bottomrule
\end{tabular}
\vspace{-0.2cm}
\label{tab:cross_dataset}
\end{table}

\begin{figure*}[t]
  \centering
  \begin{subfigure}[b]{0.2\textwidth} % Adjust the width to fit your needs
    \centering
    \includegraphics[width=\textwidth]{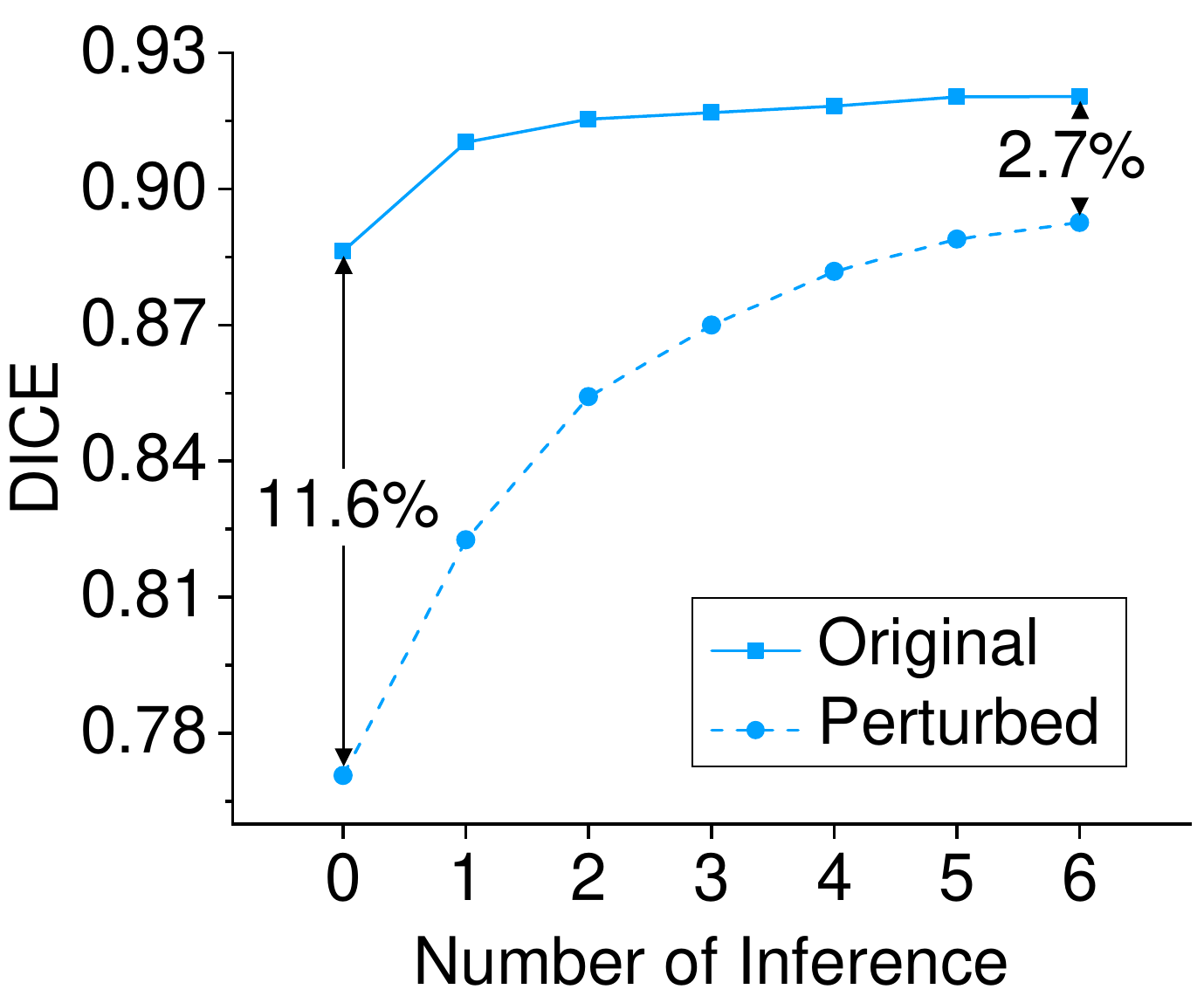} % Adjust your image file name
    \caption{Shenzhen}
    \label{fig:retrain_mask}
  \end{subfigure}
  \hspace{-0.01\textwidth}
  \begin{subfigure}[b]{0.2\textwidth} % Adjust the width to fit your needs
    \centering
    \includegraphics[width=\textwidth]{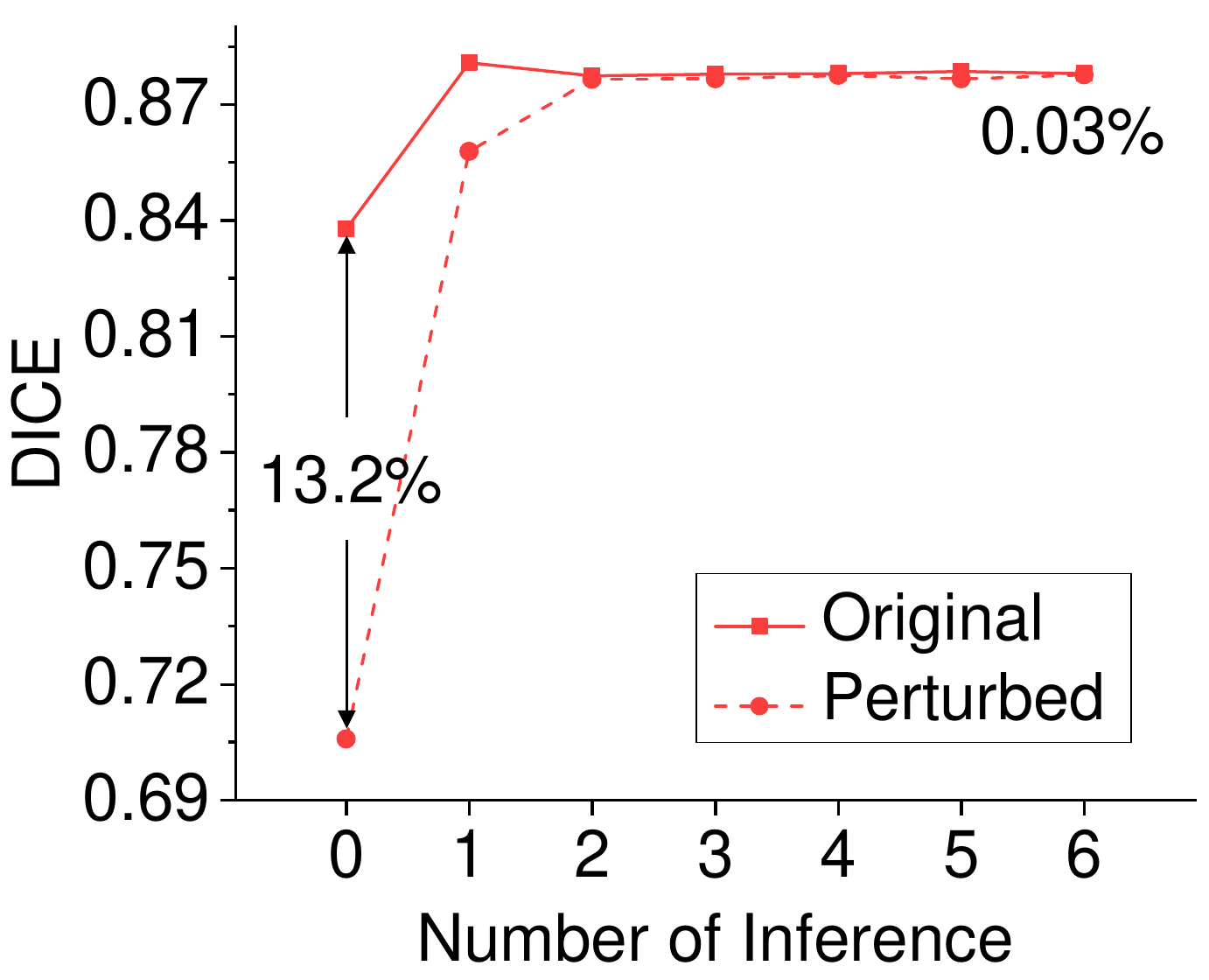} % Adjust your image file name
    \caption{OdontoAI}
    \label{fig:retrain_mask}
  \end{subfigure}
  \hspace{-0.01\textwidth}
  \begin{subfigure}[b]{0.2\textwidth} % Adjust the width to fit your needs
    \centering
    \includegraphics[width=\textwidth]{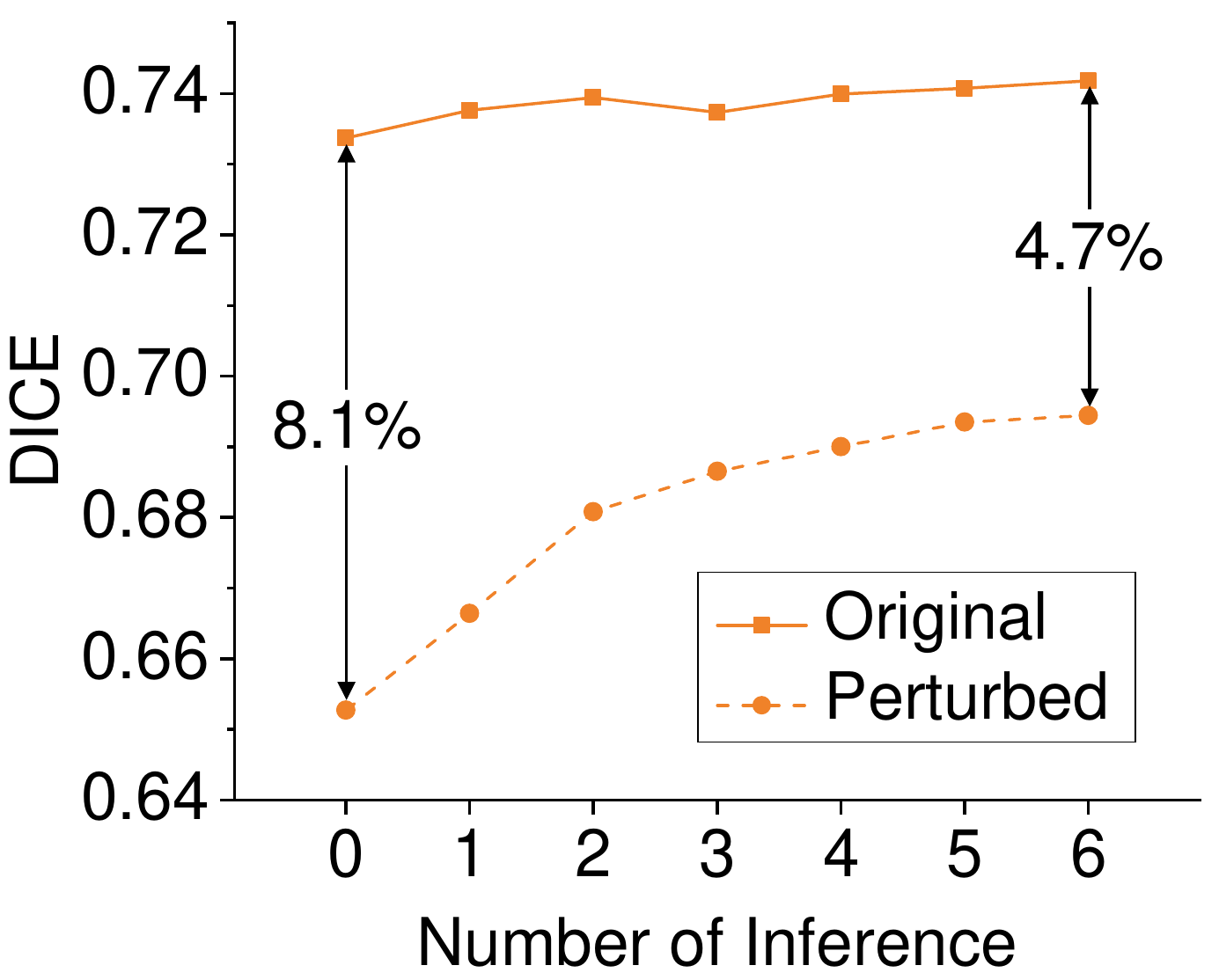} % Adjust your image file name
    \caption{CAMUS}
    \label{fig:retrain_mask}
  \end{subfigure}
  \hspace{-0.01\textwidth}
  \begin{subfigure}[b]{0.2\textwidth} % Adjust the width to fit your needs
    \centering
    \includegraphics[width=\textwidth]{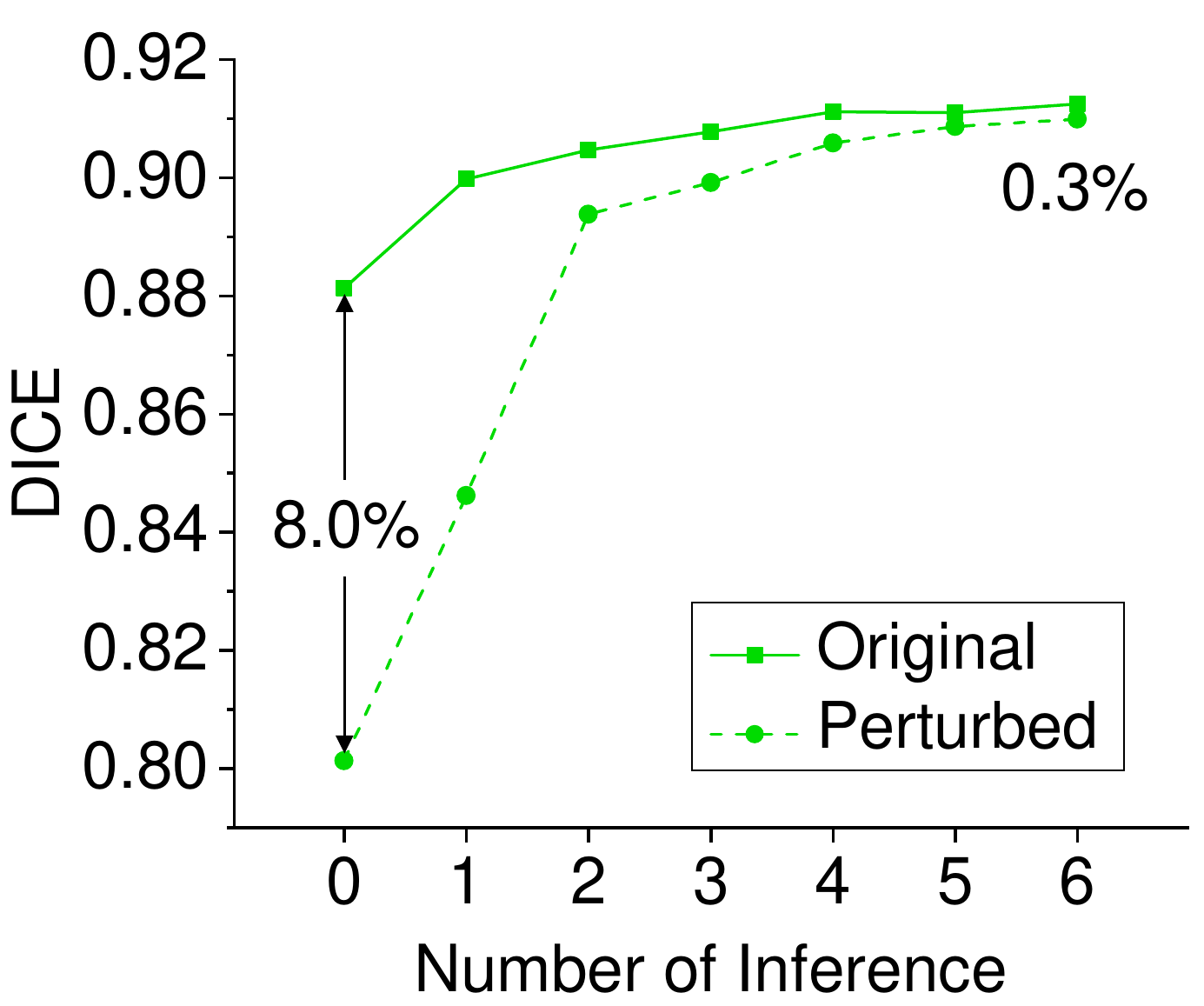} % Adjust your image file name
    \caption{JSRT}
    \label{fig:retrain_mask}
  \end{subfigure}
  \hspace{-0.01\textwidth}
  \begin{subfigure}[b]{0.2\textwidth} % Adjust the width to fit your needs
    \centering
    \includegraphics[width=\textwidth]{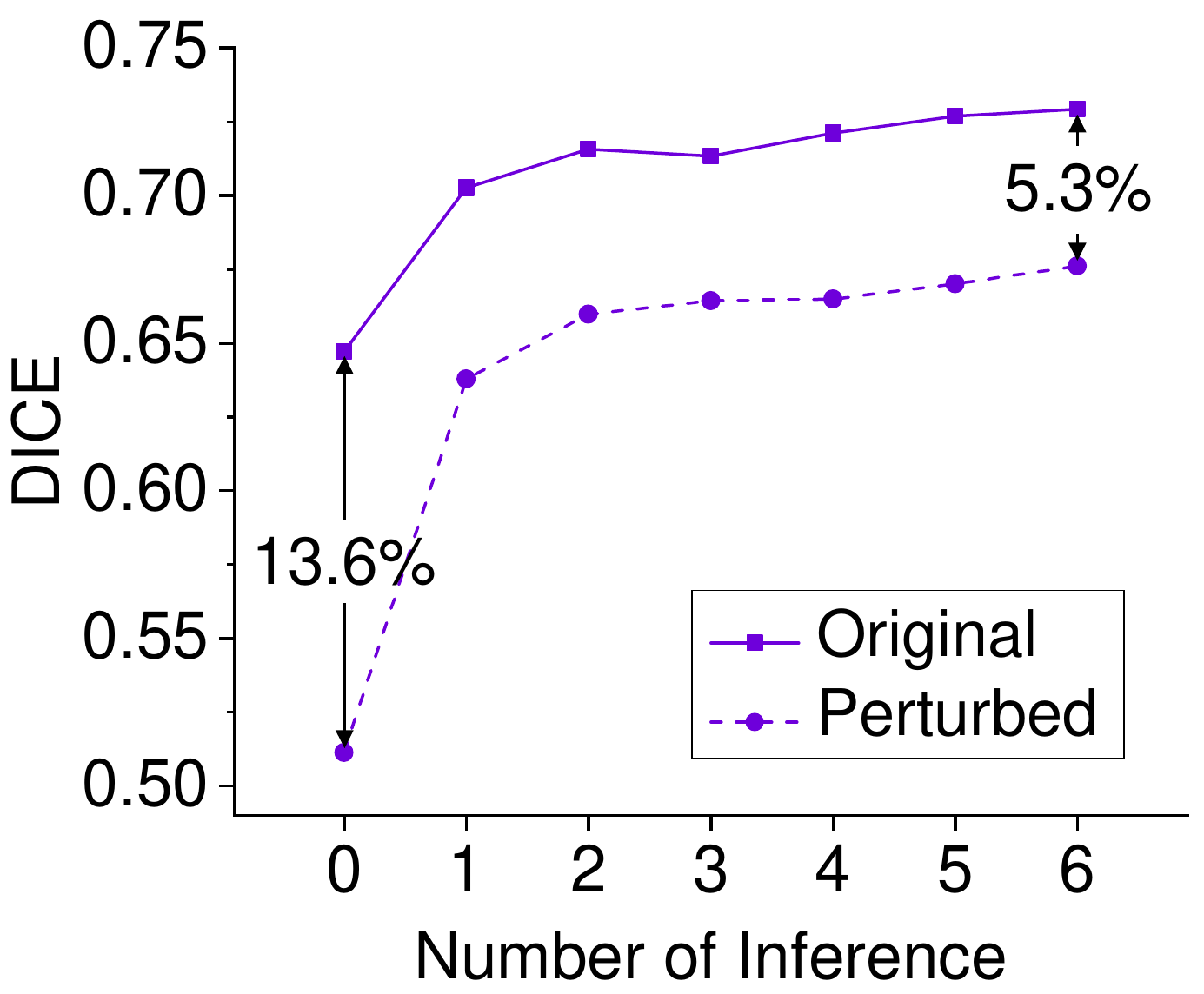} % Adjust your image file name
    \caption{BUU}
    \label{fig:retrain_mask}
  \end{subfigure}
  
  \vspace{-0.25cm}
  \caption{DICE scores of the inferences using the original and perturbed mask prompts. A count of 0 on the x-axis indicates the performance of the mask prompt generation. The numbers on the bars represent the performance gap at the beginning and the end, respectively. As the retrain number increases, the performance gets improved, with the improvement being greater in the case of the perturbed mask.}
  \label{fig:mask_perturb}
  \vspace{-0.6cm}
\end{figure*}

Additionally, we applied perturbations to the test samples (see Appendix~\ref{appendix:additional_result} for details) to assess robustness to significant changes in image characteristics (Table~\ref{tab:perturb_image}). Unlike the other top two baselines, which show significant performance drops, our model maintains relatively stable performance before and after the perturbation.

\begin{table}[h]
\footnotesize
\vspace{-0.3cm}
\caption{Change in DICE scores (\%) before and after the test sample perturbation. The value in parentheses indicates the difference. Due to space constraints, only the results with the largest differences are included. The full results are in the Appendix~\ref{appendix:additional_result}.}
\vspace{-0.2cm}
\setlength{\tabcolsep}{3pt}
\renewcommand{\arraystretch}{1}
\centering
\begin{tabular}{l|cc}
\toprule 
\textbf{Method} & \textbf{CAMUS} & \textbf{BUU} \\ 
\midrule
SegGPT & 54.9 \scalebox{0.7}[1.0]{$\rightarrow$} 45.8 \textcolor{red}{($-$9.1)} & 71.8 \scalebox{0.7}[1.0]{$\rightarrow$} 43.3 \textcolor{red}{($-$28.5)} \\
UniverSeg & 46.2 \scalebox{0.7}[1.0]{$\rightarrow$} 38.1 \textcolor{red}{($-$8.1)} & 38.4 \scalebox{0.7}[1.0]{$\rightarrow$} 32.0 \textcolor{red}{($-$6.4)}\\
\rowcolor{gray!15}  \textbf{Ours} & \textbf{74.2} \scalebox{0.7}[1.0]{$\rightarrow$} \textbf{69.2} \textcolor{red}{(\textbf{$-$5.0})} & \textbf{72.9} \scalebox{0.7}[1.0]{$\rightarrow$} \textbf{69.4} \textcolor{red}{(\textbf{$-$3.5})}\\
\bottomrule
\end{tabular}
\label{tab:perturb_image}
\vspace{-0.5cm}
\end{table}

% \vspace{-0.2cm}
% \setlength{\tabcolsep}{1.4pt}
% \renewcommand{\arraystretch}{1}
% \centering
% \begin{tabular}{lccccc}
% \toprule 
%  & \textbf{Shenzhen}& \textbf{OdontoAI} & \textbf{CAMUS} & \textbf{JSRT} & \textbf{BUU} \\ 
% \midrule
% SegGPT & 00.0 \scalebox{0.55}[1.0]{$\rightarrow$} 00.0 & 83.3 \scalebox{0.55}[1.0]{$\rightarrow$} 79.6 & 54.9 \scalebox{0.55}[1.0]{$\rightarrow$} 45.8 & 37.8 \scalebox{0.55}[1.0]{$\rightarrow$} 35.4 & 71.8 \scalebox{0.55}[1.0]{$\rightarrow$} 43.3 \\
% UniverSeg & 00.0 \scalebox{0.55}[1.0]{$\rightarrow$} 00.0 & 00.0 \scalebox{0.55}[1.0]{$\rightarrow$} 00.0 & 00.0 \scalebox{0.55}[1.0]{$\rightarrow$} 00.0 & 00.0 \scalebox{0.55}[1.0]{$\rightarrow$} 00.0 & 00.0 \scalebox{0.55}[1.0]{$\rightarrow$} 00.0 \\
% \rowcolor{gray!15}  \textbf{Ours} & 00.0 \scalebox{0.55}[1.0]{$\rightarrow$} 00.0 & 87.8 \scalebox{0.55}[1.0]{$\rightarrow$} 83.7 & 74.2 \scalebox{0.55}[1.0]{$\rightarrow$} 69.2 & 91.3 \scalebox{0.55}[1.0]{$\rightarrow$} 88.5 & 72.9 \scalebox{0.55}[1.0]{$\rightarrow$} 69.4 \\
% \bottomrule
% \end{tabular}
% \label{tab:perturb_image}
% \vspace{-0.6cm}
% \end{table}

\subsection{Perturbation of Mask Prompts}
\vspace{-0.1cm}
Although image warping between different patient images has been validated in many previous studies~\cite{balakrishnan2019voxelmorph,mansilla2020learning, jiang2024tumor, hering2022learn2reg}, and our model shows good performance across various datasets and situations, there may still be concerns regarding our method's dependence on the warped mask. To address this, we intentionally impair the initial mask prompt and measure the performance (Fig. \ref{fig:mask_perturb}). A perturbed mask prompt is created by multiplying the learned optical flow by a scalar between 0 and 1 ($\times 0.2$ in this study), reducing the performance of the initial mask prompt generation by more than 10\%. Despite this perturbation, the performance gap compared to no perturbation gets significantly reduced with repeated retraining processes. This suggests that our iterative retraining framework has a certain capacity to self-refine the predictions and find better outcomes.

\vspace{-0.1cm}
\subsection{Ablation Studies on Visual Prompts}
\vspace{-0.05cm}
Table \ref{tab:prompt_ablation}
presents the ablation results for our proposed method for determining point prompts. Similar to PerSAM \cite{zhang2023personalize}, only point and box prompts are used for comparison. The naive top-k point prompt extraction based on cosine similarity in PerSAM shows low performance. In contrast, both (1) defining candidate regions for point prompt extraction and (2) dividing them into subregions contribute to performance improvements. This suggests that inaccurate plotting due to over-reliance on pixel intensity or clustering of point prompts can degrade SAM's performance.

\begingroup
\renewcommand{\arraystretch}{1}
\begin{table}[h]
\vspace{-0.25cm}
\caption{Ablation studies on our point and box prompting strategy.}
\vspace{-0.2cm}
\scriptsize
\setlength{\tabcolsep}{0.8pt}
\centering
\begin{tabular}{l|cccccccccc}
\toprule
\multirow{2}{*}{\textbf{Method}}
 & \multicolumn{2}{c}{\textbf{Shenzhen}}&\multicolumn{2}{c}{\textbf{OdontoAI}}&\multicolumn{2}{c}{\textbf{CAMUS}}&\multicolumn{2}{c}{\textbf{JSRT}} &\multicolumn{2}{c}{\textbf{BUU}} \\ 
 & mIoU & DICE & mIoU & DICE & mIoU & DICE & mIoU & DICE & mIoU & DICE \\
\midrule
PerSAM  & 35.5 & 51.0 & 17.2 & 28.9 & 27.3 & 42.9 & 41.7 & 57.7 & 17.9 & 27.5\\
$+$ Candidate region & 54.0 & 68.5 & 50.8 & 66.0 & 37.3 & 51.4 & 76.0 & 85.1 & 43.0 & 59.2 \\ 
\rowcolor{gray!15}  $+$ Make subregions $\,$& \textbf{61.7} & \textbf{74.7} & \textbf{73.1} & \textbf{84.3} & \textbf{40.0} & \textbf{54.3} & \textbf{77.4} & \textbf{86.4} & \textbf{45.9} & \textbf{62.2} \\
\bottomrule
\end{tabular}
\label{tab:prompt_ablation}
\vspace{-0.3cm}
\end{table}
\endgroup

Table \ref{tab:ablation_prompt_type} describes the results according to the combination of different visual prompts. Using point and box prompts together has potential synergistic benefits, and adding a mask prompt further improves performance.

\begin{table}[h]
\vspace{-0.2cm}
\caption{Ablation studies for visual prompt types. As SAM does not support mask-only prompts,  the related results are excluded. P, B, and M indicate point, box, and mask prompts, respectively.}
\vspace{-0.2cm}
\footnotesize
\setlength{\tabcolsep}{1.0pt}
\renewcommand{\arraystretch}{0.97}
\centering
% \begin{threeparttable}
\begin{tabular}{l|cccccccccc}
\toprule
\multirow{2}{*}{ \textbf{Prompt} }
 & \multicolumn{2}{c}{\textbf{Shenzhen}}&\multicolumn{2}{c}{\textbf{OdontoAI}}&\multicolumn{2}{c}{\textbf{CAMUS}}&\multicolumn{2}{c}{\textbf{JSRT}} &\multicolumn{2}{c}{\textbf{BUU}} \\ 
 & mIoU & DICE & mIoU & DICE & mIoU & DICE & mIoU & DICE & mIoU & DICE \\ 
\midrule
P & 50.7 & 65.0 & 55.1 & 70.6 & 9.6 & 16.3 & 60.1 & 70.4 & 36.0 & 50.4 \\
B & 62.5 & 74.4 & 69.8 & 82.0 & 38.5 & 53.7 & 79.7 & 87.7 & 48.8 & 65.0 \\
P+B & 61.7 & 74.7 & 73.1 & 84.3 & 40.0 & 54.3 & 77.4 & 86.4 & 45.9 & 62.2 \\ 
\rowcolor{gray!15} P+B+M & \textbf{85.4} & \textbf{92.0} & \textbf{78.3} & \textbf{87.8} & \textbf{60.9} & \textbf{74.2} & \textbf{84.2} & \textbf{91.3} & \textbf{58.0 } & \textbf{72.9} \\ 
\bottomrule
\end{tabular}
\vspace{-0.3cm}
\label{tab:ablation_prompt_type}
\end{table}

Table \ref{tab:prompt_refinement} shows the results according to the number of prompt refinements. It can be observed that refining SAM's inference results at each training cycle has an effect.
\vspace{-0.2cm}

\begingroup
\renewcommand{\arraystretch}{1}
\begin{table}[h]
\caption{Effect of prompt refinement iteration number. In our main experiments, the value is set as 1.}
\vspace{-0.2cm}
\footnotesize
\setlength{\tabcolsep}{1.2pt}
\renewcommand{\arraystretch}{0.97}
\centering
\begin{tabular}{c|cccccccccc}
\toprule
\multirow{2}{*}{\textbf{Refine}}
 & \multicolumn{2}{c}{\textbf{Shenzhen}}&\multicolumn{2}{c}{\textbf{OdontoAI}}&\multicolumn{2}{c}{\textbf{CAMUS}}&\multicolumn{2}{c}{\textbf{JSRT}} &\multicolumn{2}{c}{\textbf{BUU}} \\ 
 & mIoU & DICE & mIoU & DICE & mIoU & DICE & mIoU & DICE & mIoU & DICE \\ 
\midrule
0 & 83.2  & 90.8 & 78.2 & 87.7 & 60.7 & 73.8 & \textbf{85.0} & \textbf{91.6} & 57.6 & 72.4  \\
\rowcolor{gray!15} 1 & 85.4  & 92.0 & \textbf{78.3} & \textbf{87.8} & \textbf{60.9} & \textbf{74.2} & 84.2 & 91.3 & \textbf{58.0} & \textbf{72.9} \\
2 & \textbf{86.0} & \textbf{92.4} & 77.9 & 87.5 & 59.4 & 73.0 & 81.3 & 89.4 & 54.6 & 70.0 \\ 
\bottomrule
\end{tabular}
\label{tab:prompt_refinement}
\vspace{-0.4cm}
\end{table}
\endgroup

\subsection{Other Ablation Studies}
\vspace{-0.1cm}
The ablation studies are conducted on two elements to enhance the robustness of the warping model: (1) augmentation loss and (2) affine transformation (see Eq.~\ref{eq:warp_aug} and Section~\ref{sec:implementation_detail}). Table \ref{tab:ablation_ssl} shows that the augmentation loss improves performance across all datasets, especially in the BUU dataset, which exhibits greater anatomical variation between samples. A similar trend is observed for the affine transformation (Table \ref{tab:ablation_affine}). Omitting the affine transformation and only using a deformable transformation maintains performance in other datasets but significantly reduces it for the BUU dataset. This suggests that integrating augmentation loss and affine transformation can improve robustness of warping process to large characteristic differences.

\begingroup
\renewcommand{\arraystretch}{1}
\begin{table}[h]
\vspace{-0.25cm}
\caption{Ablation results of augmentation loss.}
\vspace{-0.3cm}
\footnotesize
\setlength{\tabcolsep}{0.9pt}
\renewcommand{\arraystretch}{0.97}
\centering
\begin{tabular}{l|cccccccccc}
\toprule
\multirow{2}{*}{}
 & \multicolumn{2}{c}{\textbf{Shenzhen}}&\multicolumn{2}{c}{\textbf{OdontoAI}}&\multicolumn{2}{c}{\textbf{CAMUS}}&\multicolumn{2}{c}{\textbf{JSRT}} &\multicolumn{2}{c}{\textbf{BUU}} \\ 
 & mIoU & DICE & mIoU & DICE & mIoU  & DICE & mIoU & DICE & mIoU  & DICE \\ 
\midrule
w/o $\mathcal{L}_{\text{aug}}$ & 74.3  & 85.1 & 76.1 & 86.3 & 55.5 & 69.9 & 78.2 & 87.2 & 46.0 & 62.1  \\
\rowcolor{gray!15} w/ $\mathcal{L}_{\text{aug}}$ & \textbf{85.4} & \textbf{92.0} & \textbf{78.3} & \textbf{87.8} & \textbf{60.9} & \textbf{74.2} & \textbf{84.2} & \textbf{91.3} & \textbf{58.0} & \textbf{72.9} \\
\bottomrule
\end{tabular}
\label{tab:ablation_ssl}
\vspace{-0.3cm}
\end{table}
\endgroup

\begingroup
\renewcommand{\arraystretch}{1}
\begin{table}[h]
\vspace{-0.3cm}
\caption{Ablation results of affine transformation.}
\vspace{-0.3cm}
\footnotesize
\setlength{\tabcolsep}{0.8pt}
\renewcommand{\arraystretch}{0.97}
\centering
\begin{tabular}{l|cccccccccc}
\toprule
\multirow{2}{*}{}
 & \multicolumn{2}{c}{\textbf{Shenzhen}}&\multicolumn{2}{c}{\textbf{OdontoAI}}&\multicolumn{2}{c}{\textbf{CAMUS}}&\multicolumn{2}{c}{\textbf{JSRT}} &\multicolumn{2}{c}{\textbf{BUU}} \\ 
 & mIoU & DICE & mIoU & DICE & mIoU & DICE & mIoU & DICE & mIoU & DICE \\ 
\midrule
w/o affine & 85.2 & \textbf{92.0} & 77.4 & 87.2 & 60.4 & 73.9 & 83.9 & 91.1 & 33.7 & 49.6  \\
\rowcolor{gray!15} w/ affine & \textbf{85.4} & \textbf{92.0} & \textbf{78.3} & \textbf{87.8} & \textbf{60.9} & \textbf{74.2} & \textbf{84.2} & \textbf{91.3} & \textbf{58.0} & \textbf{72.9} \\
\bottomrule
\end{tabular}
\label{tab:ablation_affine}
\vspace{-0.5cm}
\end{table}
\endgroup

\section{Conclusion}
\vspace{-0.1cm}
We propose a novel one-shot, SAM training-free framework that generates automated visual prompts for test samples using a single reference sample in the medical domain. Our method significantly outperforms existing foundation models, including previous SAM-based studies such as PerSAM and Matcher, across various medical datasets. Without the need for additional fine-tuning of SAM or manual visual prompts, the proposed method shows robust performance under various conditions and experimental settings.

Nonetheless, our approach is not without its limitations. First, It is imperative to investigate the potential expansion of our method in few-shot learning scenarios or with 3D medical data. Additionally, our method may face challenges in certain exceptional cases where image warping is not feasible, especially when images display extreme heterogeneity due to unexpected variations in capture poses or scan ranges. We intend to leave addressing these limitations to future research. Despite these challenges, we proposed a prompt engineering method that effectively adapts SAM to the medical domain, maintaining its training-free scheme.

\clearpage
{
    \small
    \bibliographystyle{ieeenat_fullname}
    \bibliography{main}
}

\clearpage
\clearpage

\appendix

% \addcontentsline{toc}{section}{Appendix} % Add the appendix text to the document TOC
\onecolumn
\part{Appendix} 
\parttoc % Insert the appendix TOC
\twocolumn
\clearpage

\section{Dataset Details}
\label{appendix:dataset}
The experiments are conducted on five benchmark datasets that include various anatomical regions. The Shenzhen Chest X-Ray dataset \cite{jaeger2014two} was utilized as a source for lung segmentation in chest X-ray images. Specifically, we selected 566 samples from this dataset, each accompanied by bilateral lung segmentation labels. The OdontoAI dataset \cite{silva2022odontoai} comprises teeth segmentation data, and we select 555 patients presenting a full set of 32 normal teeth. Additionally, we engaged with the CAMUS dataset \cite{leclerc2019deep}, a cardiac ultrasound image dataset. We selected images from 500 patients, focusing on the 4-chamber view during the end-systolic phase. This dataset includes the left atrium and ventricle, and the left ventricular myocardium. Because the myocardium segmentation often exceeded the scan range, we limited our segmentation to the atrium and ventricle. The JSRT dataset \cite{shiraishi2000development, gaggion2022improving} was utilized for multi-class lung segmentation, where we segmented the left lung, right lung, and heart in 246 patients. Finally, the BUU dataset \cite{klinwichit2023buu} is a lumbar spine segmentation dataset. We conducted segmentation with the anterior-posterior view images from 400 patients. 

\newpage
\section{Details for the Training}
\label{appendix:warp}
\subsection{Details for Loss Function}
We primarily used SSIM loss as the basic image warping loss. However, SSIM loss can be overly sensitive to low variance regions~\cite{pambrun2015limitations}, which may lead to inefficient warping in certain cases. Therefore, for the OdontoAI and CAMUS datasets, which contain relatively large black padded areas, we used the NCC loss, which, like the SSIM function, is also a common loss function for medical image registration~\cite{meng2024correlation, balakrishnan2019voxelmorph, meng2023non}. The window sizes for calculating the SSIM and NCC losses were set to 7 and 9, respectively. 

The flow regularization term, denoted as $\mathcal{L}_{\text{reg}}$, is used to enforce smoothness in the flow field and prevent abrupt warping. It can be expressed as
\begin{align}
\mathcal{L}_{\text{reg}}(\phi_{\text{ref} \rightarrow i}) = \lVert \nabla \phi_{\text{ref} \rightarrow i} \rVert^{2}, \nonumber
\end{align}
where \(\nabla\) indicates the gradient of the given optical flow.

The segmentation loss term is defined as the sum of DICE and cross-entropy losses, with coefficients of 1 and 0.001, respectively. Additionally, the coefficient for the total segmentation loss was set to 1 in all cases.

For simplicity in notation, we omitted all coefficients in the main text. Let the coefficients for warping, regularization, and augmentation losses be denoted as $\lambda_{\text{warp}}$, $\lambda_{\text{reg}}$, and $\lambda_{\text{aug}}$, respectively. The coefficient values for each dataset are shown in Table~\ref{tab:loss_coefficient}.

\begin{table}[h]
\vspace{-0.1cm}
\caption{Loss coefficients of image-to-image warping loss in each dataset.}
\vspace{-0.2cm}
\footnotesize
\setlength{\tabcolsep}{2pt}
\renewcommand{\arraystretch}{1.1}
\centering
\begin{tabular}{c|ccccc}
\toprule
& \textbf{Shenzhen} & \textbf{OdontoAI} & \textbf{CAMUS} & \textbf{JSRT} & \textbf{BUU} \\
\midrule
$\lambda_{\text{img}}$ & 1 & 1 & 1 & 1 & 1 \\
$\lambda_{\text{reg}}$ & 0.6 & 0.9 & 0.75 & 0.75 & 0.6 \\
$\lambda_{\text{aug}}$ & 0.1 & 0.1 & 0.1 & 0.1 & 0.5 \\
\bottomrule
\end{tabular}
\label{tab:loss_coefficient}
\vspace{-0.3cm}
\end{table}

\begin{table*}[t]
\caption{Data transformation settings for each dataset. A decimal value indicates a percentage unit (e.g., 0.1 means 10\%), and the two values inside the parentheses represent the minimum and maximum values of the range, respectively. The settings for the four datasets, from the Shenzhen to the JSRT dataset, are identical.}
\footnotesize
\setlength{\tabcolsep}{2pt}
\renewcommand{\arraystretch}{1.1}
\centering
\begin{tabular}{c|ccccc}
\toprule
& \textbf{Shenzhen} & \textbf{OdontoAI} & \textbf{CAMUS} & \textbf{JSRT} & \textbf{BUU} \\
\midrule
\textbf{Rotation} & (-10, 10) & (-10, 10) & (-10, 10) & (-10, 10) & (-30, 30) \\
\textbf{Translation} & (0.0, 0.1) & (0.0, 0.1) & (0.0, 0.1) & (0.0, 0.1) & (0.1, 0.3) \\
\textbf{Scaling} & (0.9, 1.1) & (0.9, 1.1) & (0.9, 1.1) & (0.9, 1.1) & (0.8, 1.2) \\
\textbf{Shearing} & (-10, 10) & (-10, 10) & (-10, 10) & (-10, 10) & (-20, 20) \\
\midrule
\textbf{Color Jitter} & \begin{tabular}[c]{@{}c@{}} brightness: 0.2, \\ contrast: 0.2, \\ saturation: 0.2, \\ hue: 0.1 \end{tabular} & \begin{tabular}[c]{@{}c@{}} brightness: 0.2, \\ contrast: 0.2, \\ saturation: 0.2, \\ hue: 0.1 \end{tabular} & \begin{tabular}[c]{@{}c@{}} brightness: 0.2, \\ contrast: 0.2, \\ saturation: 0.2, \\ hue: 0.1 \end{tabular} & \begin{tabular}[c]{@{}c@{}} brightness: 0.2, \\ contrast: 0.2, \\ saturation: 0.2, \\ hue: 0.1 \end{tabular} & \begin{tabular}[c]{@{}c@{}} brightness: 0.4, \\ contrast: 0.4, \\ saturation: 0.4, \\ hue: 0.3 \end{tabular} \\
\midrule
\textbf{Random Crop} & \begin{tabular}[c]{@{}c@{}} crop scale: (0.9, 1.0), \\ crop ratio: (0.9, 1.1), \\ probability: 0.5 \end{tabular} & \begin{tabular}[c]{@{}c@{}} crop scale: (0.9, 1.0), \\ crop ratio: (0.9, 1.1), \\ probability: 0.5 \end{tabular} & \begin{tabular}[c]{@{}c@{}} crop scale: (0.9, 1.0), \\ crop ratio: (0.9, 1.1), \\ probability: 0.5 \end{tabular} & \begin{tabular}[c]{@{}c@{}} crop scale: (0.9, 1.0), \\ crop ratio: (0.9, 1.1), \\ probability: 0.5 \end{tabular} & \begin{tabular}[c]{@{}c@{}} crop scale: (0.8, 1.0), \\ crop ratio: (0.5, 2.0), \\ probability: 0.5 \end{tabular} \\
\bottomrule
\end{tabular}
\label{tab:aug_settings}
\vspace{-0.3cm}
\end{table*}

\subsection{Details for Augmentation}
To generate augmentation pairs, we used various transformations. We applied rotation, translation, image scaling, color jittering, and random crop. Details regarding the augmentation settings are provided in Table \ref{tab:aug_settings}. For color jittering and random crop, we name each parameter similarly to those in the torchvision package. For the four datasets excluding the BUU dataset, the augmentation settings were configured identically. For the BUU dataset, the range of augmentation was broader than that of the other datasets to enhance learning for substantial photometric and geometric variations.

\newpage 
\subsection{Details for Model Architecture}
In the warping model, we use the NICE-Trans architecture~\cite{meng2023non}. The initial encoder and decoder channels are set to 16 and 32, respectively, with the decoder containing 1 affine and 4 deformable registration layers. In the affine transformation layer, a $2 \times 3$ affine matrix is obtained as a learnable parameter, which is used to generate an affine grid for the transformation operation. The scaling ratio of the affine matrix was initialized to 1 (in the (0,0) and (1,1) elements), and the training process for these two parameters involved learning the residual values from this initial state. During the deformable registration process, the displacement for each pixel in the x and y directions is computed.

For SAM, we used SAM version 1~\cite{kirillov2023segment} to ensure a fair comparison with previous SAM-based studies~\cite{liu2023matcher,zhang2023personalize}. The image encoder backbone used was ViT-H. Additionally, we evaluated the performance of SAM2 using its largest model ($\texttt{sam2.1\_hiera\_large}$)~\cite{ravi2024sam}, which demonstrated similar performance to SAM1 (Table~\ref{tab:sam2_result}). This is mainly because the primary purpose of SAM2 is not to enhance the existing model's performance but to improve its scalability for video data. Apart from the memory attention and bank structure designed for video data processing, SAM2 is identical to SAM version 1 in terms of prompt encoding and decoding processes~\cite{ravi2024sam}. Notably, the image encoder size is actually reduced. Results from previous studies also indicate that SAM2 may perform similarly to or slightly worse than the initial version of SAM~\cite{sengupta2024sam, zhang2024unleashing}.

\begin{table}[h]
\vspace{-0.2cm}
\caption{Results of using SAM and SAM2.}
\vspace{-0.2cm}
\footnotesize
\setlength{\tabcolsep}{1.3pt}
\renewcommand{\arraystretch}{1}
\centering
\begin{tabular}{l|cccccccccc}
\toprule
\multirow{2}{*}{\textbf{Model}} & \multicolumn{2}{c}{\textbf{Shenzhen}}&\multicolumn{2}{c}{\textbf{OdontoAI}}&\multicolumn{2}{c}{\textbf{CAMUS}}&\multicolumn{2}{c}{\textbf{JSRT}} &\multicolumn{2}{c}{\textbf{BUU}} \\ 
 & mIoU & DICE & mIoU & DICE & mIoU & DICE & mIoU & DICE & mIoU & DICE \\
\midrule
SAM & 85.4 & 92.0 & 78.3 & 87.8 & 60.9 & 74.2 & 84.2 & 91.3 & 58.0 & 72.9 \\
SAM2 & 86.2 & 92.5 & 75.7 & 86.0 & 57.8 & 71.4 & 83.5 & 90.6 & 57.0 & 72.0 \\
\bottomrule
\end{tabular}
\label{tab:sam2_result}
\vspace{-0.3cm}
\end{table}

\newpage
\subsection{Details for Model Training}
For the warping model training and SAM inference, we set the batch sizes to 32 and 1, respectively. We train each warping model using the AdamW optimizer \cite{loshchilov2017decoupled} with an initial learning rate of 0.0001. During the initial warping model training phase, we trained all datasets for 30 epochs, except for the CAMUS dataset, which was trained for 20 epochs. Subsequent retraining was conducted 10 epochs per each retraining round. We limited the number of visual prompt refinement iterations to 1, and constrained the number of warping model retraining to 5. 

The majority of the experimental processes were conducted with memory requirements of 24GB. This GPU memory requirement can be flexibly adjusted by reducing the batch size during the warping model training process. The total time for one initial cycle was approximately 1 hour on a single NVIDIA V100 GPU for about 500 samples. Additionally, it took about 30 minutes to perform one cycle of retraining and re-inference.

\clearpage
\section{Other Implementation Details}
\label{appendix:experiment}
\subsection{Details for Morphological Operation}
\label{appendix:experiment_morphologic}

Regarding morphological operations, we use different kernel sizes: $7 \times 7$ for Shenzhen and JSRT, $5 \times 5$ for CAMUS, and $3 \times 3$ for OdontoAI and BUU. To quantify the number of point prompts, we employ a strategy grounded in the organ's bilaterality. For bilateral organs (e.g., bilateral lung or teeth), we designate 10 positive and negative prompts, ensuring that an equal quantity of these points is chosen from both the left and the right sides. For non-bilateral organs, we select 5 point prompts.  The morphological operations and point prompt selection were performed on a $256 \times 256$ mask. Subsequently, the results were adjusted to correspond to a $1024 \times 1024$ scale for use in the SAM. This image resolution setting was also applied consistently in the comparative experiments between PerSAM and our study to ensure a fair comparison for both models.

The backbone of our framework references the settings of PerSAM \cite{zhang2023personalize}. In the case of SAM, it can be configured to produce three distinct multi-mask outputs along with their corresponding confidence values. During the initial stage of prompt refinement, this configuration is disabled to yield a single mask output, and from the subsequent stages onwards, the one with the highest confidence from the multi-mask output is selected as the mask prompt for the next stage. Additionally, for each selected mask, boundary smoothing through erosion and dilation, as well as binary hole refilling, have been applied \cite{bradski2000opencv}. Furthermore, we adopt the ``target-semantic prompting'' strategy from PerSAM, wherein the class prototype vector is added to every input token within the image decoder to utilize the representation of the target class's visual features during SAM inference.

\newpage

\subsection{Details for Point Prompt Number Selection}
\begin{figure}[h]
    \centering
    \begin{subfigure}[h]{0.7\columnwidth}
        \centering
        \includegraphics[width=\textwidth]{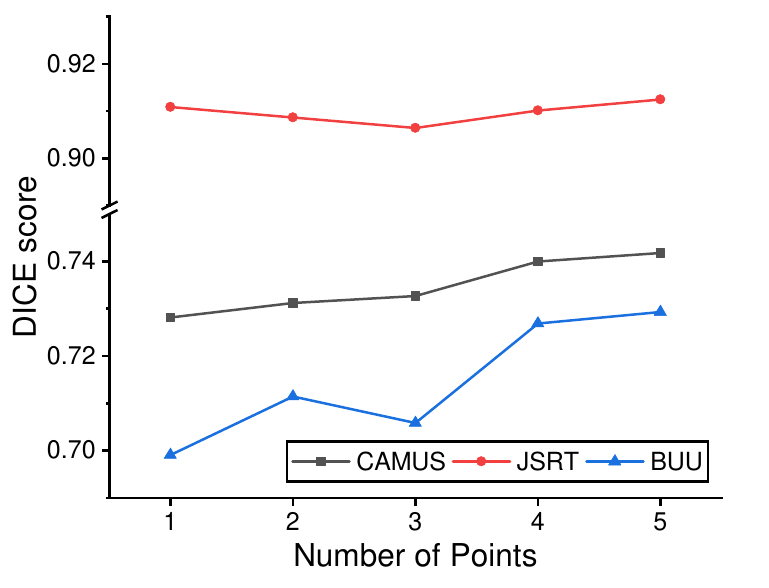}
        \caption{Results of unilateral organ datasets.}
        \label{fig:uni_point}
    \end{subfigure}\\
    \begin{subfigure}[h]{0.7\columnwidth}
        \centering
        \includegraphics[width=\textwidth]{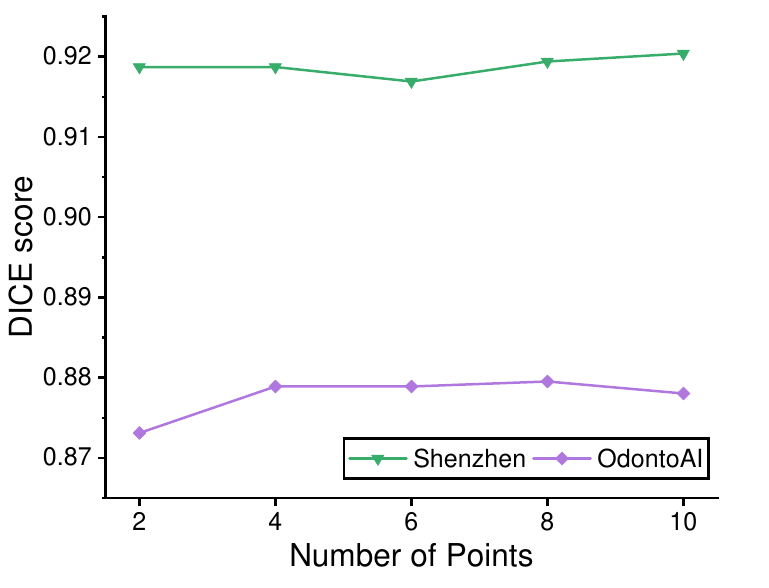}
        \caption{Results of bilateral organ datasets.}
        \label{fig:bi_point}
    \end{subfigure}
    \caption{The relationship between the number of point prompts and the DICE score in our method. We increment the number of points from 1 to 5 with a step size of 1 for unilateral organs (CAMUS, JSRT, BUU) and increase the number of points from 2 to 10 with a step size of 2 for bilateral organs (Shenzhen, OdontoAI).}
    \label{fig:num_point}
\end{figure}

We conducted evaluations by varying the number of point prompts. The results, as depicted in Figure \ref{fig:num_point}, indicate that our chosen settings—five points for unilateral organs and ten points for bilateral organs—are optimal. It is also observed that an increase in the number of prompts can lead to improvement in performance.

\begin{table*}[t]
\footnotesize
\vspace{-0.3cm}
\caption{Change in DICE scores (\%) before and after the test sample perturbation. The value in parentheses indicates the difference. The \textbf{bold} text indicates the best performance after applying the perturbation.}
\vspace{-0.2cm}
\setlength{\tabcolsep}{3.5pt}
\renewcommand{\arraystretch}{1}
\centering
\begin{tabular}{l|ccccc}
\toprule 
\textbf{Method} & \textbf{Shenzhen}& \textbf{OdontoAI} & \textbf{CAMUS}& \textbf{JSRT} & \textbf{BUU} \\ 
\midrule
SegGPT & 89.2 \scalebox{0.7}[1.0]{$\rightarrow$} 84.1 \textcolor{red}{($-$5.1)} & 83.3 \scalebox{0.7}[1.0]{$\rightarrow$} 79.6 \textcolor{red}{($-$3.7)} & 54.9 \scalebox{0.7}[1.0]{$\rightarrow$} 45.8 \textcolor{red}{($-$9.1)} & 37.8 \scalebox{0.7}[1.0]{$\rightarrow$} 35.4 \textcolor{red}{($-$2.4)} & 71.8 \scalebox{0.7}[1.0]{$\rightarrow$} 43.3 \textcolor{red}{($-$28.5)} \\
UniverSeg & 60.6 \scalebox{0.7}[1.0]{$\rightarrow$} 56.1 \textcolor{red}{($-$4.5)} & 60.1 \scalebox{0.7}[1.0]{$\rightarrow$} 44.4 \textcolor{red}{($-$15.7)} & 46.2 \scalebox{0.7}[1.0]{$\rightarrow$} 38.1 \textcolor{red}{($-$8.1)} & 75.7 \scalebox{0.7}[1.0]{$\rightarrow$} 61.0 \textcolor{red}{($-$14.7)} & 38.4 \scalebox{0.7}[1.0]{$\rightarrow$} 32.0 \textcolor{red}{($-$6.4)}\\
\rowcolor{gray!15}  \textbf{Ours} & \textbf{92.0} \scalebox{0.7}[1.0]{$\rightarrow$} \textbf{87.8} \textcolor{red}{\textbf{($-$4.2)}} & \textbf{87.8} \scalebox{0.7}[1.0]{$\rightarrow$} \textbf{83.7} \textcolor{red}{\textbf{($-$4.1})} & \textbf{74.2} \scalebox{0.7}[1.0]{$\rightarrow$} \textbf{69.2} \textcolor{red}{(\textbf{$-$5.0})} & \textbf{91.3} \scalebox{0.7}[1.0]{$\rightarrow$} \textbf{88.5} \textcolor{red}{\textbf{($-$2.8})} & \textbf{72.9} \scalebox{0.7}[1.0]{$\rightarrow$} \textbf{69.4} \textcolor{red}{(\textbf{$-$3.5})}\\
\bottomrule
\end{tabular}
\label{tab:perturb_image_full}
\vspace{-0.2cm}
\end{table*}

\newpage
\section{Additional Experimental Results}
\label{appendix:additional_result}
\subsection{Perturbation to Test Samples}
For the experiments described in Section~\ref{sec:image_robustness}, the default augmentation settings applied to the test images and corresponding masks are summarized in Table~\ref{tab:perturb_detail}. Random crop was excluded as it can cause excessive changes to the segmentation label. To introduce types of perturbations that were not included in our augmentation loss, the following two transformations were newly applied: Gaussian noise addition and Gaussian blur. For the Shenzhen dataset, all other settings remained the same, but the standard deviation of the Gaussian noise was adjusted to 0.5 to create significant performance degradation across all models. Additionally, Gaussian blur was omitted for this dataset. For model training, all the settings described in Appendix~\ref{appendix:warp} and Appendix~\ref{appendix:experiment} were maintained, except for reducing the initial training epochs from 30 to 10 for the lung dataset to prevent excessive overfitting to images with added Gaussian noise, which was of higher intensity compared to other datasets. 

\begin{table}[h]
\vspace{-0.2cm}
\caption{Default settings for test sample perturbation.}
\vspace{-0.2cm}
\footnotesize
\renewcommand{\arraystretch}{1.1}
\centering
\begin{tabular}{c|c}
\toprule
\textbf{Transformation} & \textbf{Range} \\
\midrule
\textbf{Rotation} & (-20, 20) \\
\textbf{Translation} & (0.0, 0.1) \\
\textbf{Scaling} & (0.8, 1.2) \\
\textbf{Shearing} & (-20, 20) \\
\midrule
\textbf{Color Jitter} & \begin{tabular}[c]{@{}c@{}} brightness: 0.3, \\ contrast: 0.3, \\ saturation: 0.3, \\ hue: 0.2 \end{tabular} \\
\midrule
\textbf{Gaussian Noise} & \begin{tabular}[c]{@{}c@{}} standard deviation: 0.1, \\ probability: 0.5 \end{tabular}  \\ 
\midrule
\textbf{Gaussian Blur} & \begin{tabular}[c]{@{}c@{}} kernel size: 7, \\ SD range: (0.1, 2.0), \\ 
probability: 0.5 \end{tabular} \\
\bottomrule
\end{tabular}
\label{tab:perturb_detail}
\vspace{-0.2cm}
\end{table}

The inference results for perturbed samples across all datasets are presented in Table~\ref{tab:perturb_image_full}. For comparison, we also evaluated SegGPT and UniverSeg, the top two baseline models, using the same experimental setup. While our model demonstrates stable performance across all datasets before and after perturbation, the baseline models often exhibits significant performance degradation, particularly in the CAMUS and BUU datasets, which display greater sample heterogeneity than the other datasets.

\clearpage
\section{Qualitative Results}
\label{appendix:qual}
This section presents visualized qualitative examples from the main experiments and the ablation studies. We compare the results of point prompting between our model and two SAM-based baseline methods, PerSAM and Matcher (Section \ref{appendix:qual_point_prompt_persam}). In addition, we demonstrate the outcomes of iteratively refining visual prompts and retraining the warping model (Section \ref{appendix:qual_examples_refine}). Finally, we include additional qualitative results for the main experiment (Section \ref{appendix:qual_examples_add}).

\subsection{Failure Cases of Point Prompting in PerSAM and Matcher}
\label{appendix:qual_point_prompt_persam}

\begin{figure*}[h]
    \centering
    \begin{subfigure}[h]{0.95\textwidth}
        \centering
        \includegraphics[width=\textwidth]{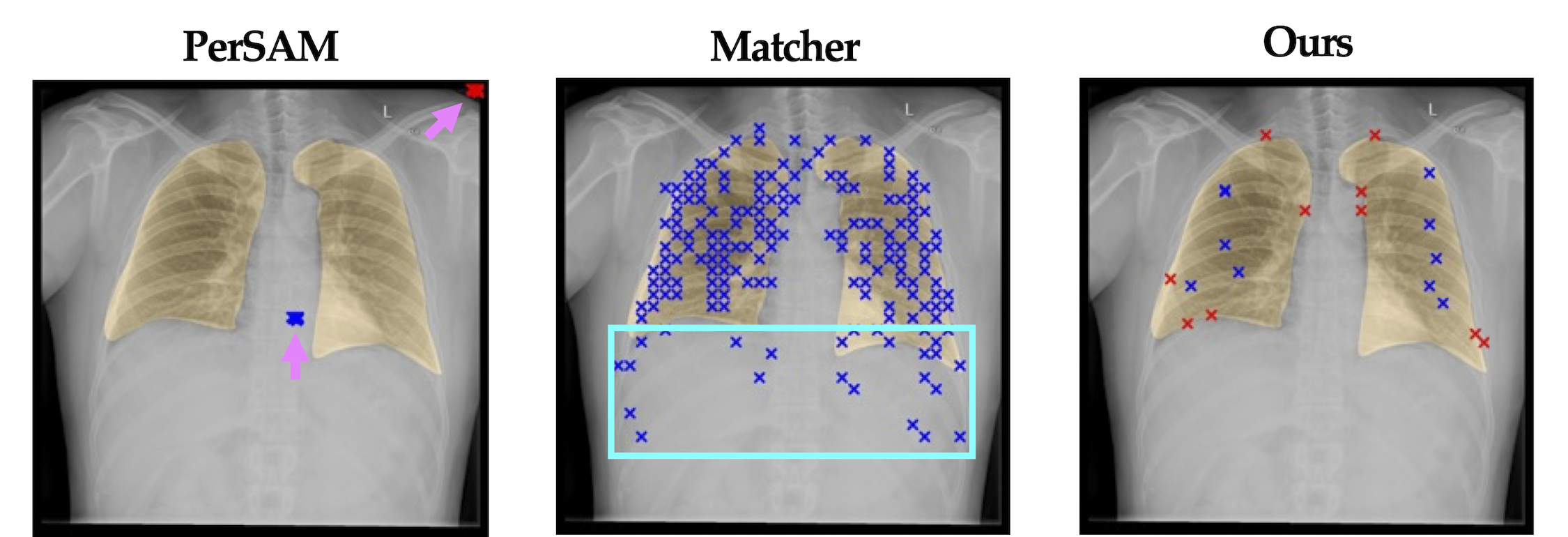}
        \caption{Example from the Shenzhen dataset.}
        \label{fig:persam_prompt_lung}
    \end{subfigure} \\
    \begin{subfigure}[h]{0.95\textwidth}
        \centering
        \includegraphics[width=\textwidth]{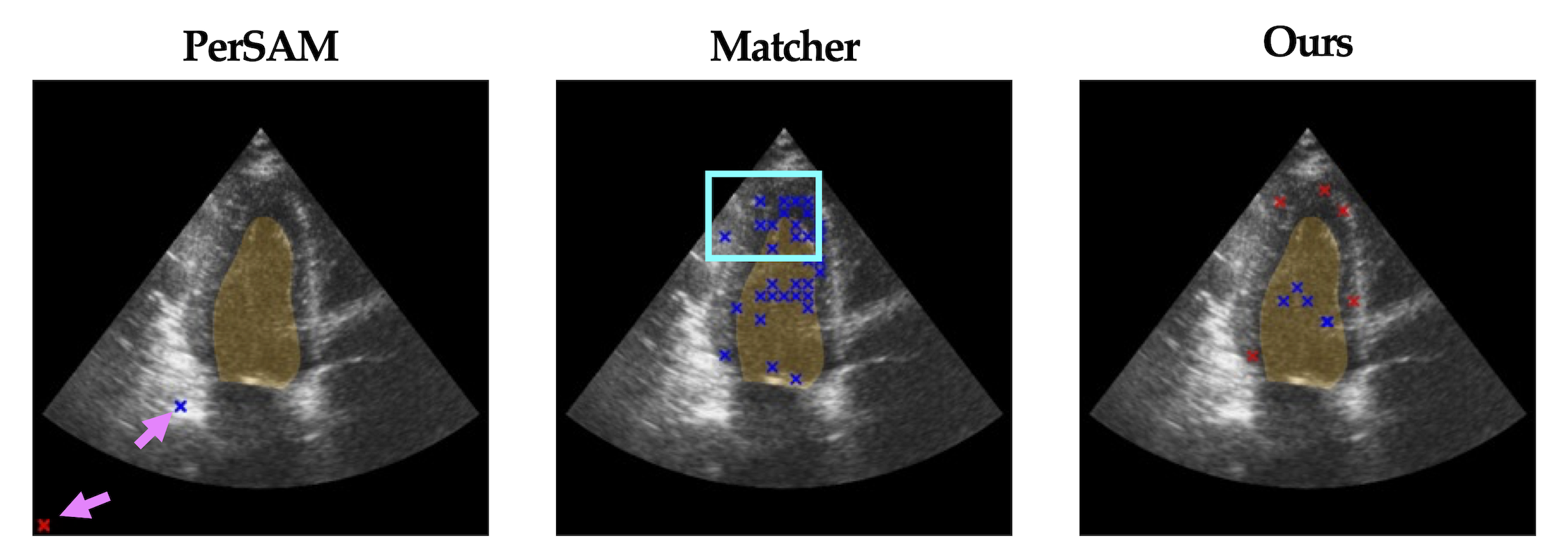}
        \caption{Example from the CAMUS dataset.}
        \label{fig:persam_prompt_cardiac} 
    \end{subfigure} \\
    \begin{subfigure}[h]{0.95\textwidth}
        \centering
        \includegraphics[width=\textwidth]{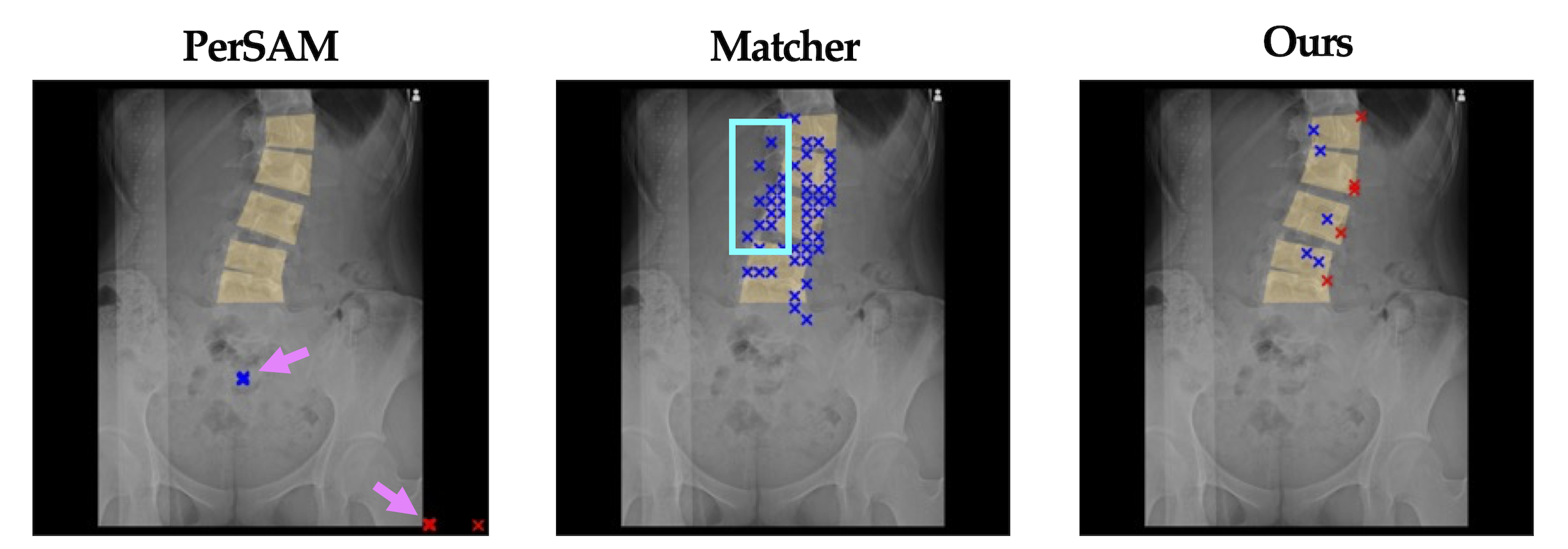}
        \caption{Example from the BUU dataset.}
        \label{fig:persam_prompt_spine} 
    \end{subfigure}
    \caption{Comparative examples of point prompts from PerSAM \cite{zhang2023personalize}, Matcher \cite{liu2023matcher}, and our approach in Shenzhen, CAMUS, and BUU datasets.  The ground truth masks are highlighted in yellow, with positive and negative point prompts marked in blue and red, respectively. Similar to Figure~\ref{fig:persam_prompt}, PerSAM and Matcher exhibit over-clustering (\textcolor{magenta}{pink arrows}) or struggle to differentiate organs with similar pixel intensities (\textcolor{cyan}{skyblue boxes}).}
    \label{fig:persam_prompt_add}
\end{figure*}

Figure \ref{fig:persam_prompt_add} presents additional results of point prompting across three SAM-based models—PerSAM, Matcher, and our proposed model. Consistent with the results shown in Figure~\ref{fig:persam_prompt}, both SAM-based baseline models exhibit over-clustering (indicated by pink arrows) or misprompting to surrounding points with intensities resembling those of the target organs (highlighted by sky blue boxes). Specifically, PerSAM demonstrates a noticeable clustering of both positive and negative prompts: positive points frequently group around non-target organs with comparable intensity, while negative points tend to concentrate in the black-padded periphery. Matcher also struggles to differentiate organs with similar intensity ranges, with the absence of a negative prompt being a significant limitation. In contrast, the point prompts in our method align more accurately with the ground truth masks while avoiding the clustering issue.

\subsection{Examples of Prompt Refinement and Retraining}
\label{appendix:qual_examples_refine}
\begin{figure*}[h]
    \captionsetup{font=small}
    \centering
    \begin{subfigure}[h]{0.95\textwidth}
        \centering
        \captionsetup{font=small}
        \includegraphics[width=\textwidth]{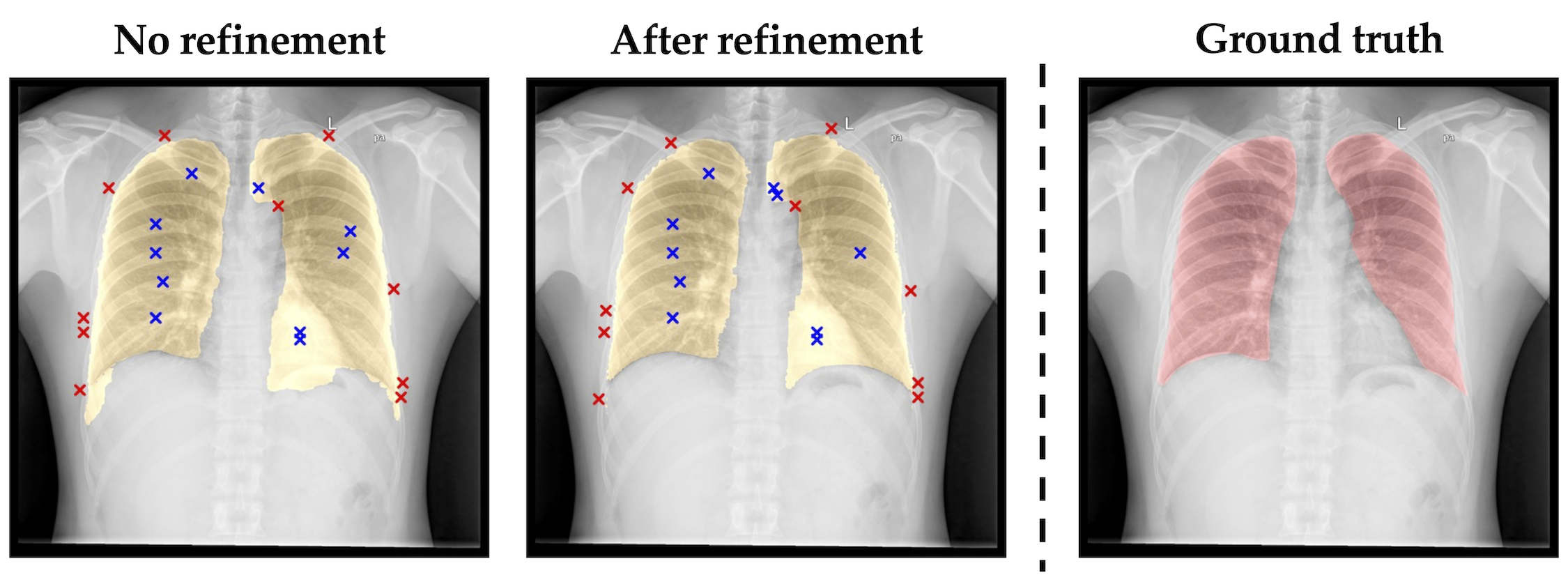}
        \caption{Example from the Shenzhen dataset.}
        \label{fig:iterative_refine_lung}
    \end{subfigure} \\
    \begin{subfigure}[h]{0.95\textwidth}
        \centering
        \captionsetup{font=small}
        \includegraphics[width=\textwidth]{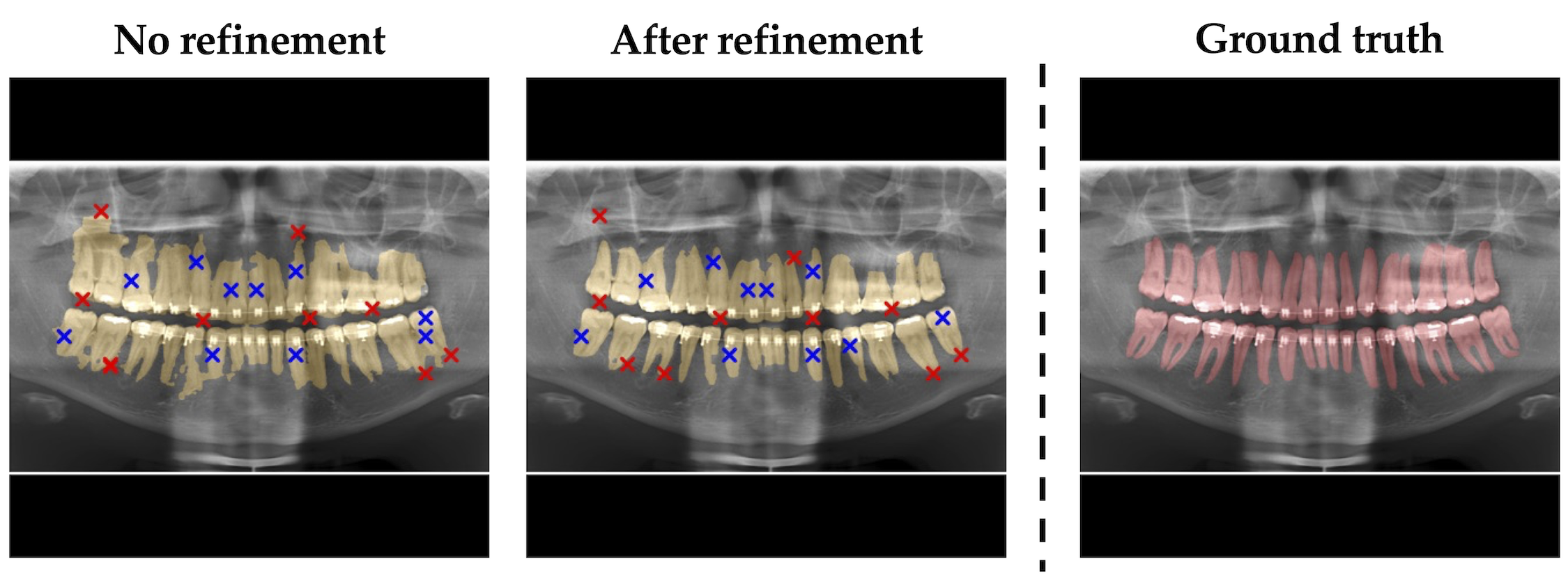}
        \caption{Example from the OdontoAI dataset.}
        \label{fig:iterative_refine_teeth}
    \end{subfigure} \\
    \begin{subfigure}[h]{0.95\textwidth}
        \centering
        \captionsetup{font=small}
        \includegraphics[width=\textwidth]{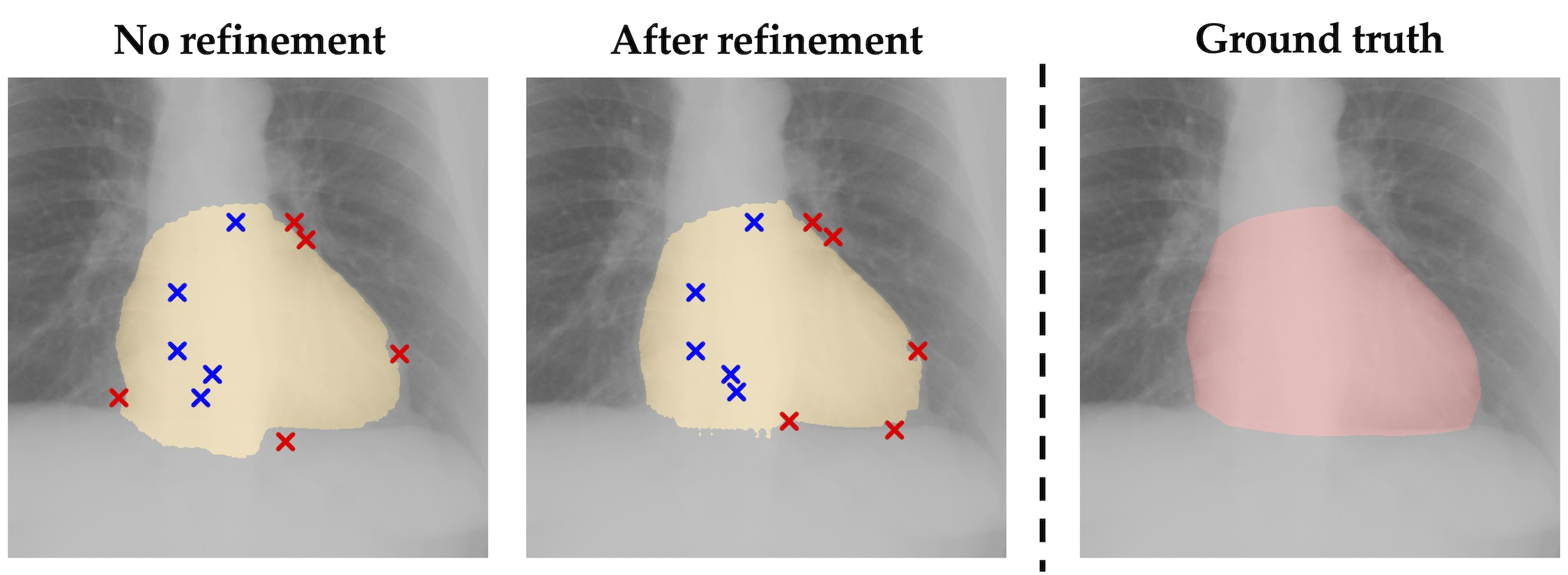}
        \caption{An example of heart segmentation from the JSRT dataset.}
        \label{fig:iterative_refine_add_jsrt}
    \end{subfigure}
    \caption{Additional examples of iterative prompt refinement. The predicted mask output is indicated in yellow, while the positive and negative point prompts are marked in blue and red, respectively.}
    \label{fig:iterative_refine_add}
\end{figure*}

Figure \ref{fig:iterative_refine_add} presents additional results of iterative visual prompt refinement. Similar to Figure \ref{fig:iterative_refine}, it demonstrates that as the refinement process progresses, the mask prompt is adjusted, and the offsets of the point prompts are dynamically updated. The performance improvements achieved through this visual prompt refinement are consistent with the findings and trends observed in PerSAM.

\begin{figure*}[h]
    \centering
    \captionsetup{font=small}
    \begin{subfigure}[h]{0.95\textwidth}
        \centering
        \captionsetup{font=small}
        \includegraphics[width=\textwidth]{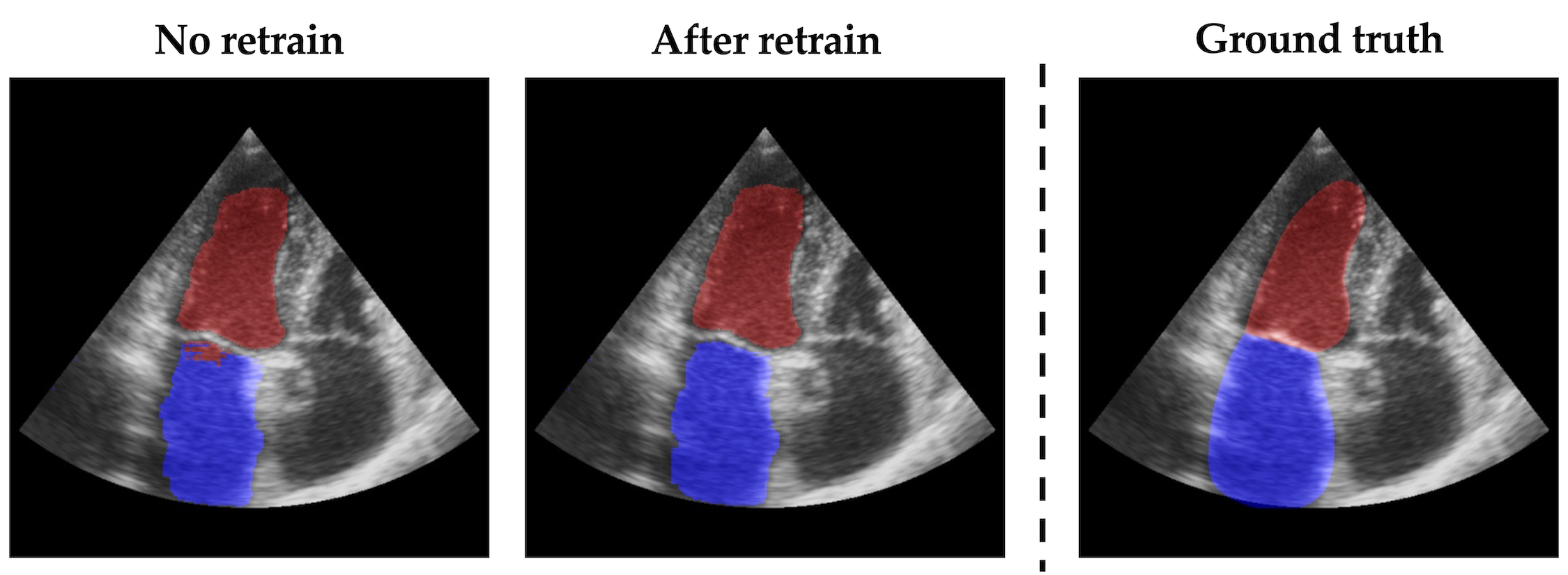}
        \caption{Example from the CAMUS dataset.}
        \label{fig:iterative_retrain_cardiac}
    \end{subfigure} \\
    \begin{subfigure}[h]{0.95\textwidth}
        \centering
        \captionsetup{font=small}
        \includegraphics[width=\textwidth]{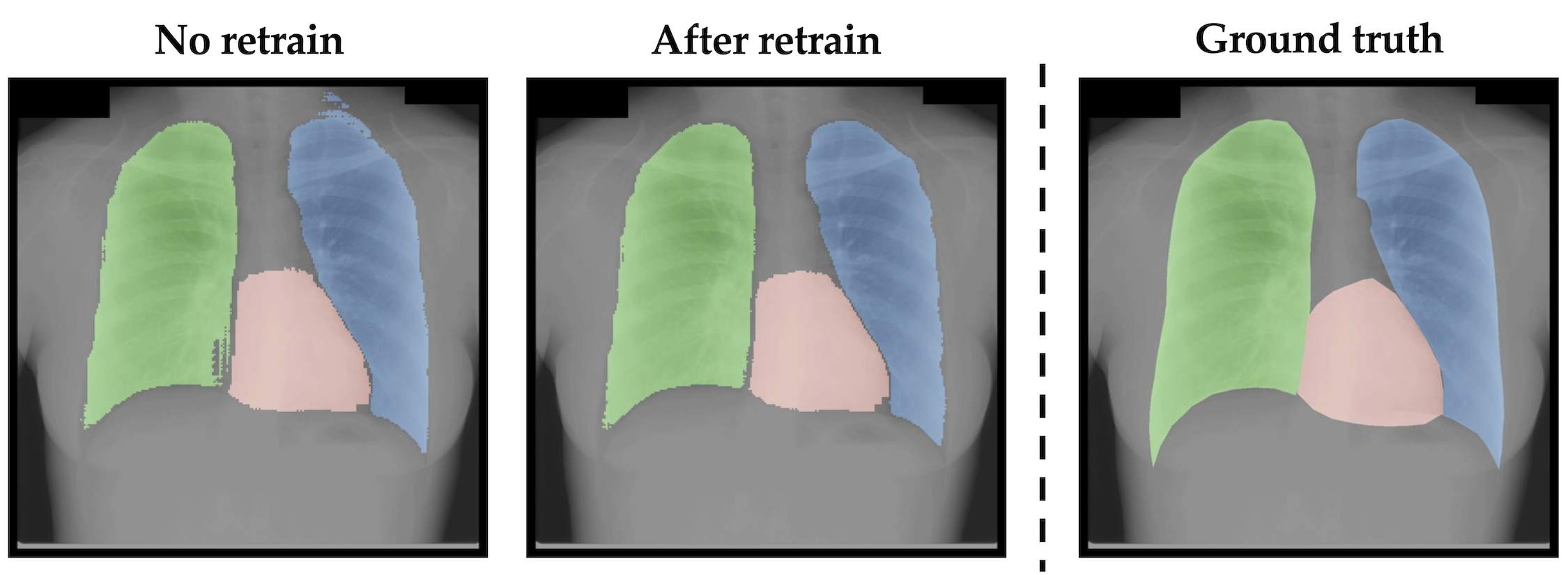}
        \caption{Example from the JSRT dataset.}
        \label{fig:iterative_retrain_jsrt}
    \end{subfigure} \\
    \begin{subfigure}[h]{0.95\textwidth}
        \centering
        \captionsetup{font=small}
        \includegraphics[width=\textwidth]{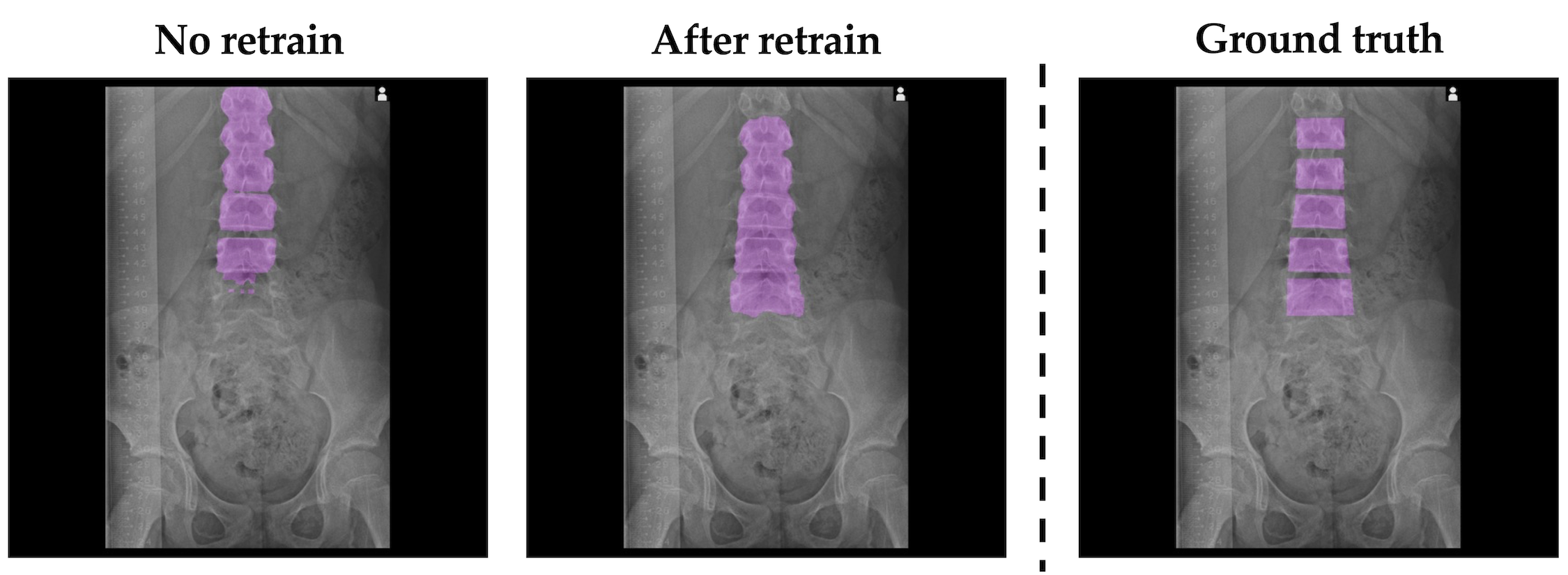}
        \caption{Example from the BUU dataset.}
        \label{fig:iterative_retrain_buu}
    \end{subfigure} \\
    \caption{
Illustrative examples of the iterative warping model retraining process. The results are presented for (1) the model without retraining and (2) the model after a maximum of five retraining iterations.}
    \label{fig:iterative_retrain_add}
\end{figure*}

Figure \ref{fig:iterative_retrain_add} presents qualitative results before and after retraining the warping model. In line with the trends shown in Figure~\ref{fig:mask_perturb}, the output segmentation quality improves with additional retraining, which leverages the inference results from the previous training iteration as pseudolabels. Combined with visual prompt refinement, this retraining strategy equips our proposed framework with a self-correction capacity for improving outcomes.

\newpage
\subsection{Additional Segmentation Results}
\label{appendix:qual_examples_add}
\begin{figure*}[b]
  \centering
    \includegraphics[width=\textwidth]{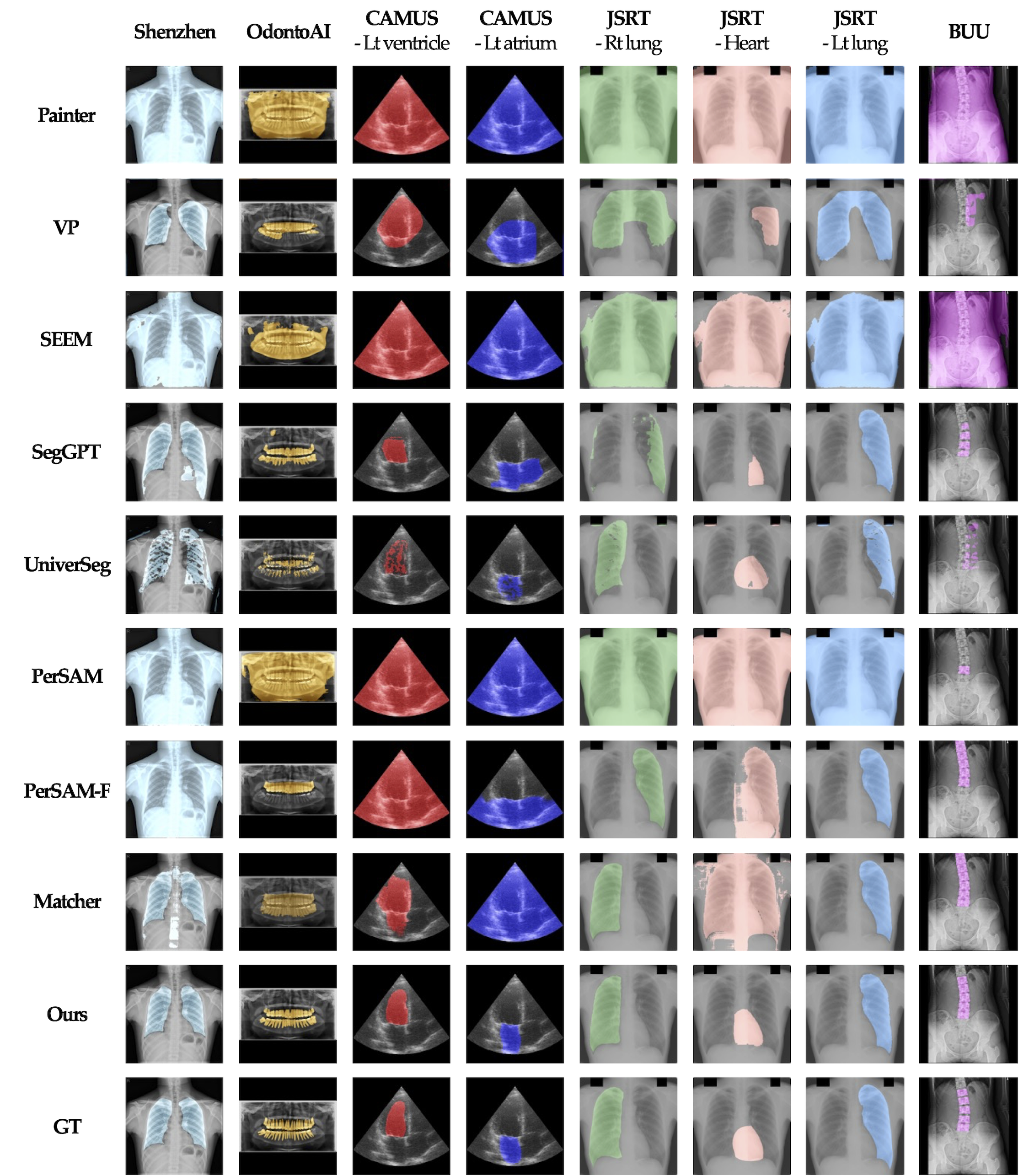}
  \vspace{-0.15cm}
  \caption{Further qualitative analysis of baseline and our proposed models.}
  \label{fig:qual_analysis_add1}
\end{figure*}

\begin{figure*}[h]
  \centering
    \includegraphics[width=\textwidth]{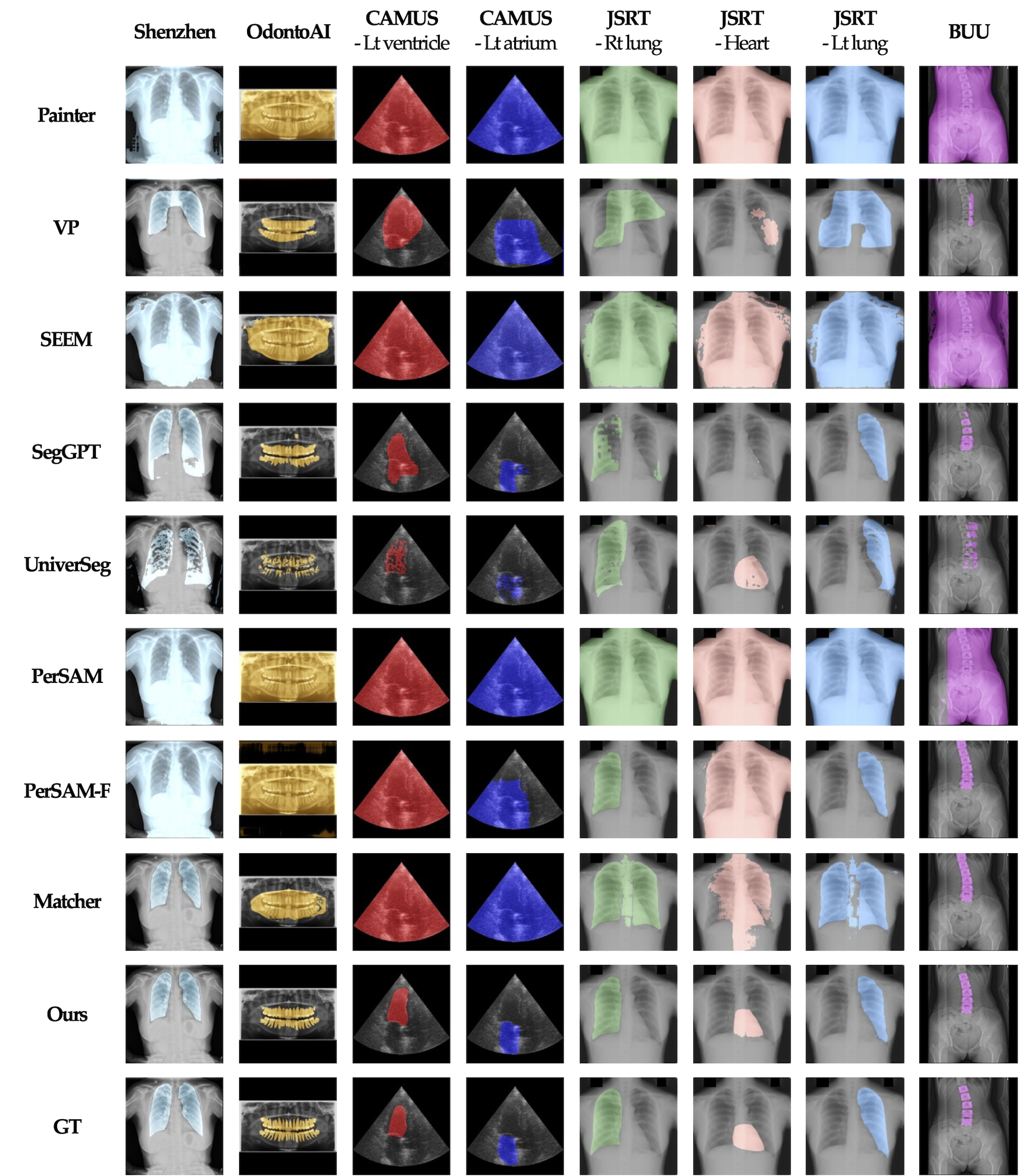}
  \vspace{-0.15cm}
  \caption{Another qualitative analysis of baseline and our proposed models.}
  \label{fig:qual_analysis_add2}
\end{figure*}

Figures \ref{fig:qual_analysis_add1} and \ref{fig:qual_analysis_add2} provide additional qualitative examples of model predictions. Similar to Figure \ref{fig:qual_analysis}, the baseline models often fail to accurately capture the subsegment, leading to segmentation at excessively large scales or incorrectly segmenting other organs with similar intensity to the target organ. In contrast, our method consistently demonstrates stable performance across all datasets.

\begin{figure*}[h]
    \centering
    \captionsetup{font=small}
    \begin{subfigure}[h]{0.95\textwidth}
        \centering
        \captionsetup{font=small}
        \includegraphics[width=\textwidth]{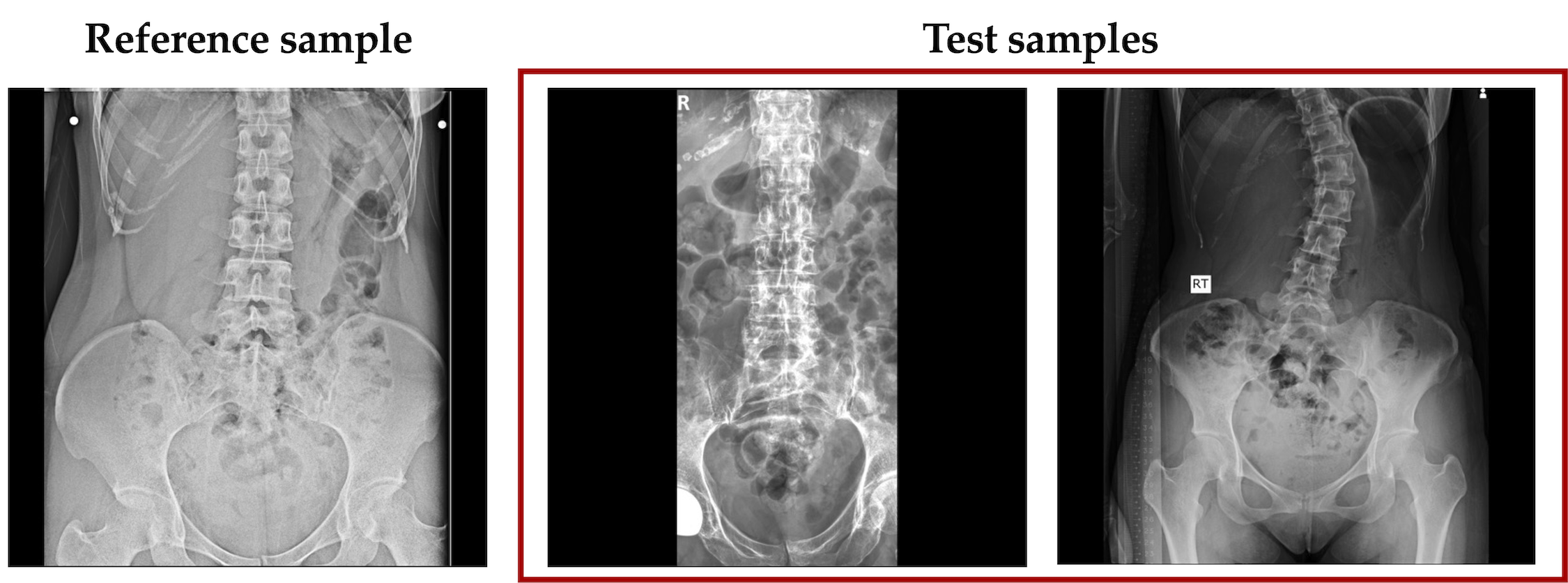}
        \caption{Reference image and test images that exhibit significant characteristic differences.}
        \label{fig:spine_hard_image}
    \end{subfigure} \\
    \begin{subfigure}[h]{0.95\textwidth}
        \centering
        \captionsetup{font=small}
        \includegraphics[width=\textwidth]{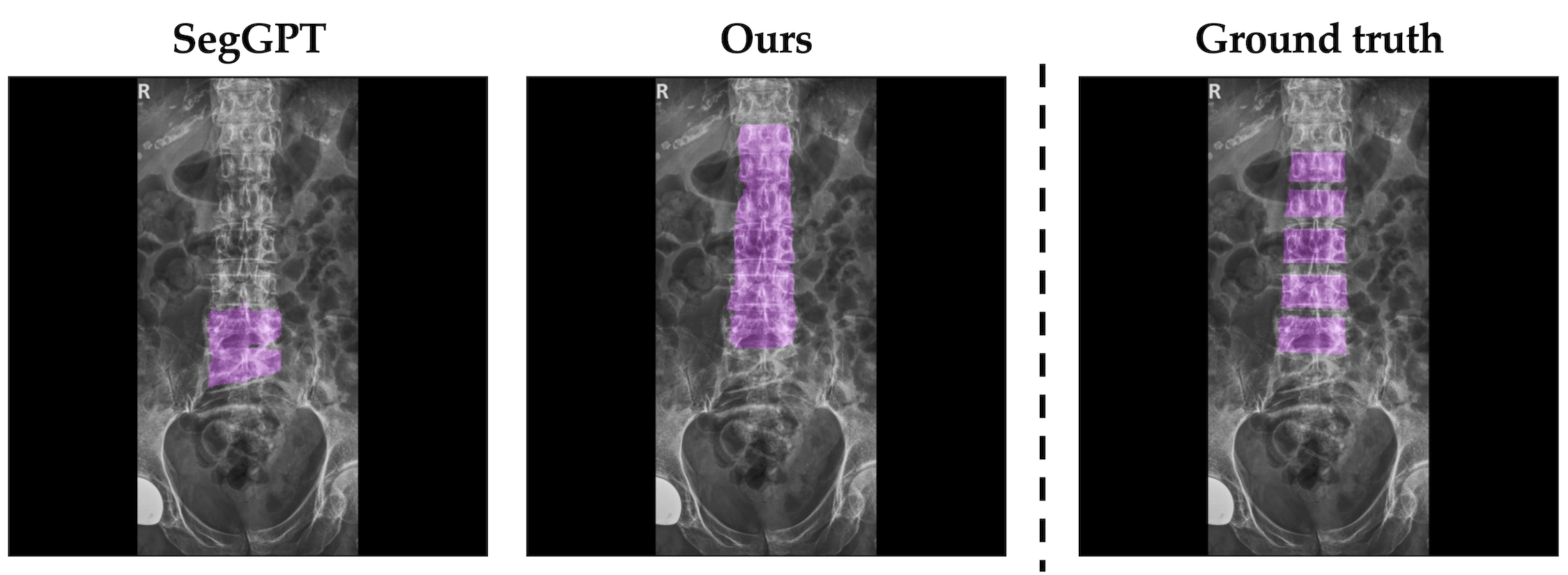}
        \caption{Inference results of the first test sample.}
        \label{fig:spine_hard_image_output1}
    \end{subfigure} \\
    \begin{subfigure}[h]{0.95\textwidth}
        \centering
        \captionsetup{font=small}
        \includegraphics[width=\textwidth]{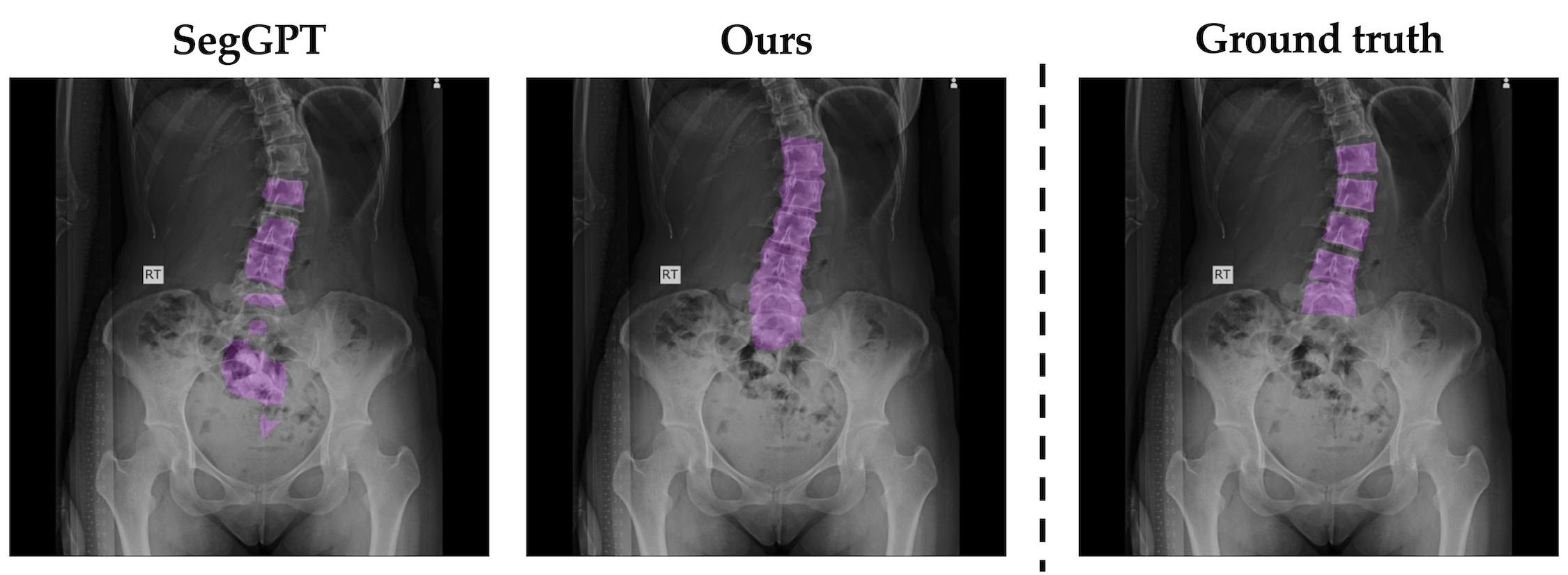}
        \caption{Inference results of the second test sample.}
        \label{fig:spine_hard_image_output2}
    \end{subfigure} \\
    \caption{
Inference results of SegGPT and our model, with examples of large differences in test samples taken from the BUU dataset. (a) Compared to the reference image, the test samples exhibit significant differences. 
(b) and (c) show the corresponding segmentation results for the two test samples presented in (a).}
    \label{fig:spine_hard}
\end{figure*}

The qualitative results for the large photometric and geometric difference gap between the reference image and the test image are demonstrated in Figure~\ref{fig:spine_hard}. We visualized the results using samples from the BUU dataset, which demonstrates greater variation in sample characteristics compared to other datasets. The results of SegGPT, the baseline model with the best performance, are presented alongside our results. Despite the significant deviation of the test samples from the reference sample, as shown in Figure~\ref{fig:spine_hard_image}, our model demonstrates relatively robust performance compared to SegGPT (see Figs.~\ref{fig:spine_hard_image_output1} and~\ref{fig:spine_hard_image_output2}).  

% \clearpage

% \clearpage
% \small
% \setlength{\bibsep}{0pt}
% \bibliographystyle{ieeenat_fullname}
% \bibliography{supplementary}

\end{document}